\documentclass[lettersize,journal]{IEEEtran}
\usepackage{amsmath,amsfonts}
\usepackage{algorithmic}
\usepackage{algorithm}
\usepackage{array}
\usepackage[caption=false,font=normalsize,labelfont=sf,textfont=sf,farskip=0pt]{subfig}
\usepackage{textcomp}
\usepackage{stfloats}
\usepackage{url}
\usepackage{verbatim}
\usepackage{graphicx}
\usepackage{cite}
\usepackage{mathtools}
\usepackage{times}
\usepackage{epsfig}
\usepackage{graphicx}
\usepackage{amsmath}
\usepackage{amssymb}

\usepackage{multirow}
\usepackage{tabularx}
\usepackage{booktabs}
\usepackage{longtable}
\usepackage{makecell}

\usepackage[accsupp]{axessibility}  
\usepackage[font=footnotesize,labelfont=bf,belowskip=-5pt,aboveskip=3pt]{caption} 

\usepackage{dsfont}
\usepackage{pifont}
\usepackage{url}
\usepackage{color}
\usepackage[dvipsnames]{xcolor}
\usepackage[export]{adjustbox}
\usepackage{comment}
\usepackage{algorithm} 
\usepackage{multirow}
\usepackage{diagbox}

\usepackage[utf8]{inputenc} 
\usepackage[T1]{fontenc}    
\usepackage{url}            
\usepackage{amsfonts}       
\usepackage{nicefrac}       
\usepackage{microtype}      
\usepackage{enumitem}
\usepackage{blindtext}
\usepackage{sidecap}
\usepackage{wrapfig}
\definecolor{citecolor}{RGB}{65,105,225}
\usepackage[pagebackref=true,breaklinks=true,colorlinks,bookmarks=false,citecolor=citecolor]{hyperref}

\usepackage{dsfont}

\definecolor{dg}{rgb}{0,0.694,0.298}
\definecolor{purple}{rgb}{0.4,0.176,0.569}
\definecolor{royalblue}{RGB}{65,105,225}
\usepackage{pifont}
\newcommand{\cmark}{\textcolor{dg}{\ding{52}}}%
\newcommand{\xmark}{\textcolor{red}{\ding{56}}}%
\newcommand{\figref}[1]{Fig.~\ref{#1}}
\newcommand{\reqref}[1]{Eq.~(\ref{#1})}
\newcommand{\secref}[1]{Sec.~\ref{#1}}
\newcommand{\tableref}[1]{Tab.~\ref{#1}}

\makeatletter
\DeclareRobustCommand\onedot{\futurelet\@let@token\@onedot}
\def\@onedot{\ifx\@let@token.\else.\null\fi\xspace}
\def\eg{\emph{e.g}\onedot} 
\def\ie{\emph{i.e}\onedot} 
 
\def\etc{\emph{etc}\onedot}

\makeatother

\definecolor{americanrose}{rgb}{1.0, 0.01, 0.24}

\definecolor{mygray}{RGB}{200,200,200}
\definecolor{mygreen}{RGB}{0,176,80}
\definecolor{myyellow}{RGB}{255,192,0}
\definecolor{mypink}{RGB}{255,64,255}
\definecolor{mycyan}{RGB}{0,255,255}

\makeatletter
\begin{document}

\title{Time-variant Image Inpainting via Interactive Distribution Transition Estimation}


\author{Yun Xing$^*$, Qing Guo$^{\dagger}$,~\IEEEmembership{Senior Member,~IEEE,} Xiaoguang Li, Yihao Huang, Xiaofeng Cao,~\IEEEmembership{Member,~IEEE,} Di Lin,~\IEEEmembership{Member,~IEEE,} Ivor Tsang,~\IEEEmembership{Fellow,~IEEE,} Lei Ma,~\IEEEmembership{Member,~IEEE,}
\thanks{$^*$ This work was done during intership at IHPC and CFAR, A*STAR, Singapore. $^\dagger$ Qing Guo is the corresponding author.}
\thanks{Yun Xing is with the University of Alberta, Edmonton, AB T6G 2R3, Canada (e-mail: yxing8@ualberta.ca).}
\thanks{Qing Guo and Ivor Tsang are with IHPC and CFAR, A*STAR, Singapore. (e-mail: tsingqguo@ieee.org and ivor tsang@cfar.a-star.edu.sg)}
\thanks{Xiaoguang Li is with University of South Carolina, USA. (e-mail: xl22@email.sc.edu)}
\thanks{Yihao Huang is with Nanyang Technological University, Singapore. (e-mail: huangyihao22@gmail.com)}
\thanks{Xiaofeng Cao is with the School of Artificial Intelligence, Engineering Research Center of Knowledge-Driven Human-Machine Intelligence, Ministry of Education, Jilin University, Changchun, Jilin 130012, China.
(e-mail: xiaofengcao@jlu.edu.cn)}
\thanks{Di Lin is with the College of Intelligence and Computing, Tianjin University, Tianjin 300072, China. (e-mail: ande.lin1988@gmail.com)}
\thanks{Lei Ma is with the University of Tokyo, Tokyo 113-0033, Japan, and also with the University of Alberta, Edmonton, AB T6G 2R3, Canada. (e-mail: ma.lei@acm.org)}}

\markboth{Journal of \LaTeX\ Class Files,~Vol.~14, No.~8, August~2021}%
{Shell \MakeLowercase{\textit{et al.}}: A Sample Article Using IEEEtran.cls for IEEE Journals}


\maketitle

\begin{abstract}
In this work, we focus on a novel and practical task, \ie, Time-vAriant iMage inPainting (TAMP). 
The aim of TAMP is to restore a damaged target image by leveraging the complementary information from a reference image, where both images captured the same scene but with a significant time gap in between, \ie, time-variant images.
%
%
Different from conventional reference-guided image inpainting, the reference image under TAMP setup presents significant content distinction to the target image and potentially also suffers from damages.
%
%
Such an application frequently happens in our daily lives to restore a damaged image by referring to another reference image, where there is no guarantee of the reference image's source and quality.
In particular, our study finds that even state-of-the-art (SOTA) reference-guided image inpainting methods fail to achieve plausible results due to the chaotic image complementation.
%
To address such an ill-posed problem, we propose a novel Interactive Distribution Transition Estimation (InDiTE) module which interactively complements the time-variant images with adaptive semantics thus facilitate the restoration of damaged regions.
%
%
To further boost the performance, we propose our TAMP solution, namely Interactive Distribution Transition Estimation-driven Diffusion (InDiTE-Diff), which integrates InDiTE with SOTA diffusion model and conducts latent cross-reference during sampling.
%
%
Moreover, considering the lack of benchmarks for TAMP task, we newly assembled a dataset, \ie, TAMP-Street, based on existing image and mask datasets.
We conduct experiments on the TAMP-Street datasets under two different time-variant image inpainting settings, which show our method consistently outperform SOTA reference-guided image inpainting methods for solving TAMP.
%
%
\end{abstract}


\section{Introduction}
\label{sec:intro}

\begin{figure}
    \centering
    \includegraphics[width=\linewidth]{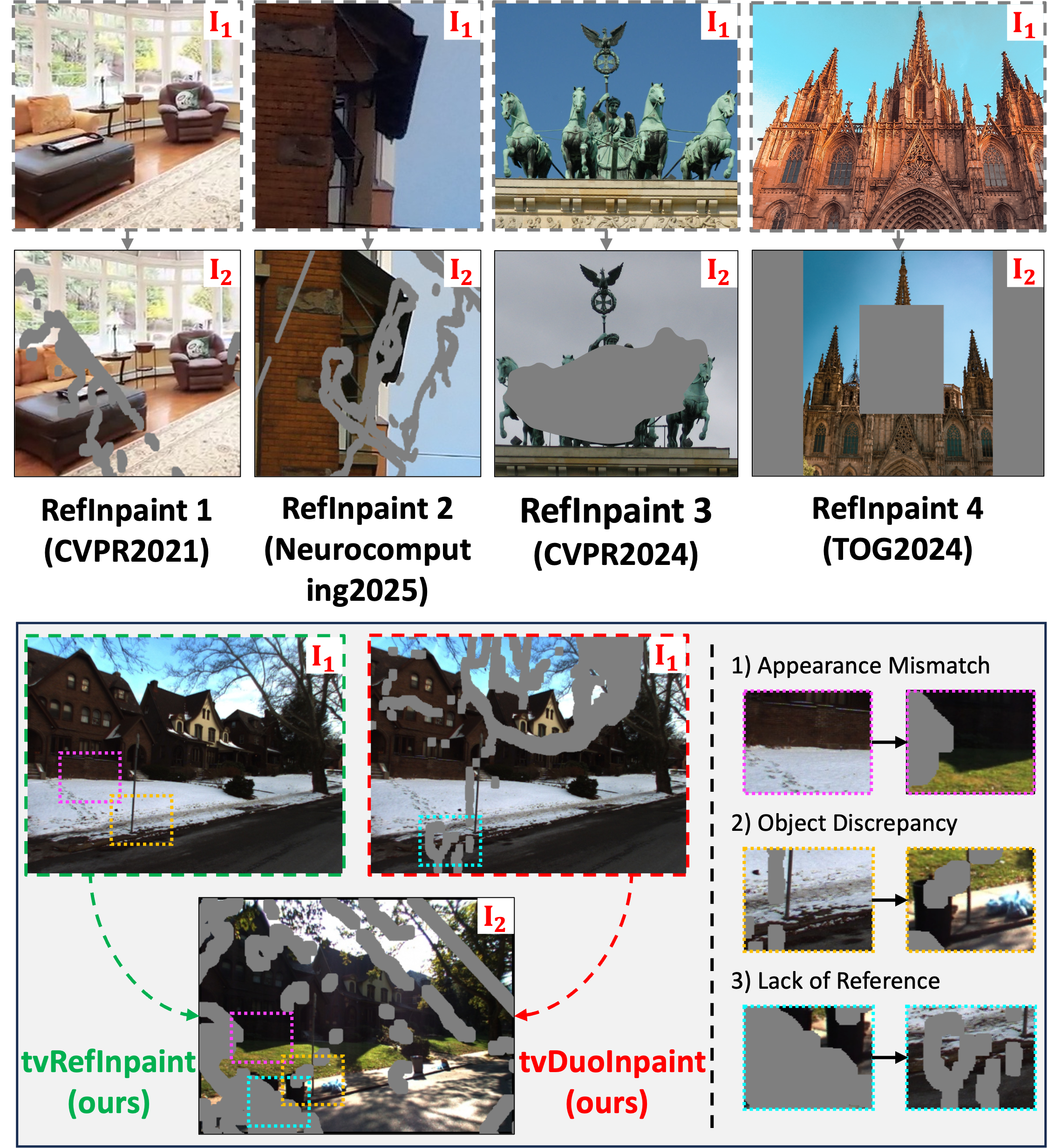}
    \caption{Illustration of the reference and target image pairs $(\textcolor{red}{\mathbf{I}_1},\textcolor{red}{\mathbf{I}_2})$ for \textit{(upper)} existing RefInpaint studies and \textit{(lower)} our time-variant image inpainting setting. \colorbox{mygray}{Gray regions} represent the image damages that need to be inpainted.}
    \label{fig:existing_test}
\end{figure}

\IEEEPARstart{I}{mage} inpainting \cite{lugmayr2022repaint} \cite{zhang2023copaint} \cite{zhang2024reference} \cite{wang2021dynamic} refers to the task of restoring an image suffering from pixel missing (\ie, the target image) to its original content.
At present, traditional image inpainting methods \cite{guo2021jpgnet} \cite{li2022misf} \cite{wang2023coarse} still struggle to recover the target image when the damage is extensive, especially for those regions with complex semantic structures.
To tackle such ``ill-posed'' problem, reference-guided image inpainting \cite{oh2019onion}, abbreviated as RefInpaint, was proposed where another image of the same scene, \ie, the reference image, is introduced and serves as the inpainting guidance.
By involving extra prior knowledge, reference-guided image inpainting methods \cite{zhou2021transfill} \cite{wang2024rego} \cite{liao2023transref} \cite{cao2024leftrefill} have achieved promising results for more practical application scenarios.
However, existing reference-guided image inpainting studies usually assume the reference image to be intact and has consistent object and appearance to the target image (see \figref{fig:existing_test} \emph{top}), both of which fail to align with the real-world situations thus limit the application of RefInpaint techniques.
%

Essentially, existing RefInpaint studies suppose we can easily find a reference image, which captured the same scene to the target damaged image, with the help of online search.
While such collection is feasible, it is impractical to assume the image quality and content consistency of the reference image are exactly of desire for inpainting purpose.
On the one hand, it is common that the reference image is captured with a large time gap to the target image which potentially leads to object and appearance divergence.
As the \textcolor{mygreen}{\textbf{tvRefInpaint}} (time-variant reference-guided image inpainting) exemplified in the \textit{bottom} of \figref{fig:existing_test}, it may be at winter when capturing the scene for reference (\ie, $\mathbf{I}_1$) while the damaged image $\mathbf{I}_2$ is taken in summer.
Obviously, either object or appearance of the damaged regions are significantly changed, \ie, 1) appearance mismatch and 2) object discrepancy highlighted in \textcolor{mypink}{pink} and \textcolor{myyellow}{yellow} rectangles respectively.
%
%
Moreover, the reference image at disposal may comes from historical source which potentially also suffers from pixel missing.
As the \textcolor{red}{\textbf{tvDuoInpaint}} (time-variant duo-image inpainting) in the \textit{bottom} of \figref{fig:existing_test} shows, there may be no reference at all for some damaged regions when the reference image is also damaged, \ie, 3) Lack of Reference in \textcolor{mycyan}{cyan} rectangle.
It is clear that both the two situations are out of existing RefInpaint works' scope where the image constraint is exposed only to provide explicit reference.
In fact, \textbf{\textit{the more practical expectation is that we can access reference images of the same scene but there potentially exists significant temporal change with no image quality guarantee.}}
%

%
%
%
%
%
%

\begin{figure*}[ht]
    \centering
    \includegraphics[width=\linewidth]{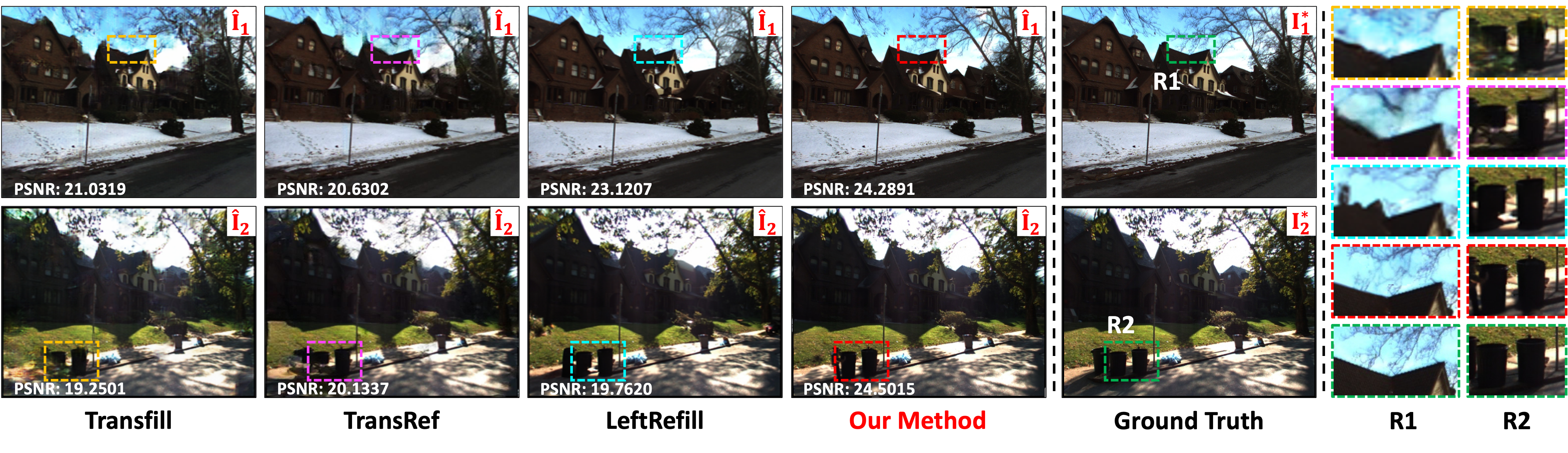}
    \caption{Visualization of image inpainting results with existing RefInpaint methods and our method for the tvDuoInpaint case in \figref{fig:existing_test}.}
    \label{fig:test_compare}
\end{figure*}

However, refill the damaged regions with original contents under such setting is a non-trivial task.
Distinguished from conventional image inpainting challenges (\eg, parallax), either appearance mismatch or object discrepancy are ill-posed as there is no exact reference to facilitate the restoration.
Not to mention there are regions with no reference at all when the reference image also suffers from damage.
Yet, it is still doable for us attempting to inpaint the time-variant images with existing RefInpaint models.
As shown in \figref{fig:test_compare}, we evaluate existing RefInpaint methods with representative time-variant images.
It can be seen that even the state-of-the-art (SOTA) generative method (\ie, LeftRefill \cite{cao2024leftrefill}) cannot achieve realistic results.
We attribute such inpainting failure to the inappropriate image complementation where the semantic correspondences are debilitated by the object and appearance divergence (see empirical analysis in \secref{subsec:motivation}).
To effectively leverage the available reference contents, in this work, we propose to consider the complementation between the time-variant images as a distribution transition estimation process based on the fact that they share the near-identical overall geometry.
%

In specific, we explicitly designed a module, named Interactive Distribution Transition Estimation (InDiTE), to learn the transition process.
By interactively merging features and predicatively filtering semantics \cite{li2022misf} during the learning of the transition, InDiTE complements the time-variant images with consistent semantics and identifies the inappropriate contents that should be suppressed. 
Practically, these functions are realized with two specialized heads where one responsible for assembling the complementation results and another predicting the confidence of the complementation.
Such a designing further facilitate us to address the lack of reference challenge where we utilize generative model to fill the low-confidence regions that indicating poor complemented regions due to the lack of reference.
Specifically, we build on SOTA diffusion model (\ie, DDNM \cite{wang2022ddnm}) and propose our TAMP solution, dubbed Interactive Distribution Transition Estimation-driven Diffusion (InDiTE-Diff), which further strength the consistency between time-variant images with cross-reference during the diffusion sampling.
In summary, our study contributes to the image inpainting research with following aspects
\begin{itemize}[leftmargin=0.5cm]
    \item We present the Time-vAriant iMage inPainting (TAMP) task which further enhances the practical applications of existing reference-guided image inpainting research by considering the factual existence of the large time gap between the reference and target images, namely time-variant images.
    \item To address the TAMP task, we first propose to consider the complementation between the time-variant images as a distribution transition process and design a novel module, Interactive Distribution Transition Estimation (InDiTE), to learn such process for appropriate complementation.
    \item Benefit to the plug-and-play nature of InDiTE, we then further build on SOTA diffusion model and propose the Interactive Distribution Transition-driven Diffusion (InDiTE-Diff) as our final TAMP solution.
    \item Due to lack of benchmarks, we build the TAMP-Street dataset to evaluate existing and our propose methods for solving TAMP. The comprehensive experiments conducted on TAMP-Street demonstrate the proposed InDiTE and InDiTE-Diff enjoys great superiority over existing methods.
\end{itemize}

\section{Related Work}
\label{sec:relate}

As a long-standing image restoration task, there are plenty of research studies \cite{quan2024deep} focus on the problem of image inpainting. Here, we briefly review some related works for reference.

\textbf{Traditional image inpainting} mainly focus on the designing of different inpainting pipelines, \ie, single-shot, two-stage and progressive method. 
The single-shot methods essentially learn a mapping from a corrupted image to the completed one, \eg, mask-aware design \cite{zhu2021image} \cite{wang2021parallel} \cite{wang2021dynamic}, attention mechanism \cite{zhang2022w} \cite{he2022masked} \cite{zheng2022bridging} and many others \cite{zeng2022aggregated} \cite{lu2022glama} \cite{feng2022generative} \cite{wang2022dual}.
As for two-stages methods, coarse-to-fine \cite{kim2022zoom} \cite{roy2021image} and structure-then-texture \cite{yamashita2022boundary} \cite{wu2021deep} are two main adopted methodologies.
Some further works \cite{zeng2020high} \cite{li2022srinpaintor} extend the two-stages method and propose to inpaint image progressively, where the image holes are iteratively completed from the boundary to the center.
Parallel to the pipeline designing, generative model is also popular for image inpainting research.
Researchers initially utilize VAE \cite{kingma2013auto} and GAN \cite{goodfellow2014generative} as the backbone to generate missing contents for the corrupted image \cite{zheng2021pluralistic, zheng2022image}. 
%
%
Later, some researches also adopt flow-based methods \cite{wang2022diverse} and masked language models \cite{wan2021high} for image inpainting tasks.
Most recently, diffusion models \cite{ho2020denoising} with dedicated sampling strategy designing \cite{lugmayr2022repaint} \cite{zhang2023copaint} \cite{wang2022ddnm} have become the main utilized generative model for image inpainting and have achieved superior inpainting performance.

\textbf{Reference-guided image inpainting} was recently proposed to tackle the failing of traditional methods when the holes are large or the expected contents have complicated semantic layout.
With extra reference images introduced, those methods \cite{zhou2021transfill} \cite{liu2022reference} \cite{li2022reference} \cite{liao2023transref} \cite{cao2024leftrefill} manage to achieve more visually convincing inpainting results.
Specifically, \cite{zhou2021transfill} propose a multi-homography fusion pipeline combined with deep warping, color harmonization, and single image inpainting to address the issue of parallax between the target and reference images.
Later \cite{liu2022reference} proposes to separately infer the texture and structure features of the input image considering their pattern discrepancy of texture and structure during inpainting.
\cite{liao2023transref} introduced a transformer-based encoder-decoder model to better harmonize the style differences, where they constructed a publicly accessible benchmark dataset containing 50K pairs of images.
More recently, \cite{cao2024leftrefill} adopts the powerful Visual Language Model (VLM) for further boosting the performance and has demonstrate the VLM potential for the reference-guided image inpainting task.

%

\section{Time-variant Image Inpainting}
\label{sec:problem}

In order to discuss time-variant image inpainting with concreteness, we first formulate the problem and highlight the challenges in Sec.~\ref{subsec:formulate}.
Next in Sec.~\ref{subsec:motivation}, we resort to SOTA reference-guided image inpainting method (\ie, LeftRefill \cite{cao2024leftrefill}) for an empirical study.
By analyzing the results, we determine the root cause of the poor performance when applying existing method to solve time-variant image inpainting task, which motivates us to propose our solution (\ie, InDiTE) in Sec.~\ref{sec:method}.

\begin{figure}[t]
    \centering
    \includegraphics[width=\linewidth]{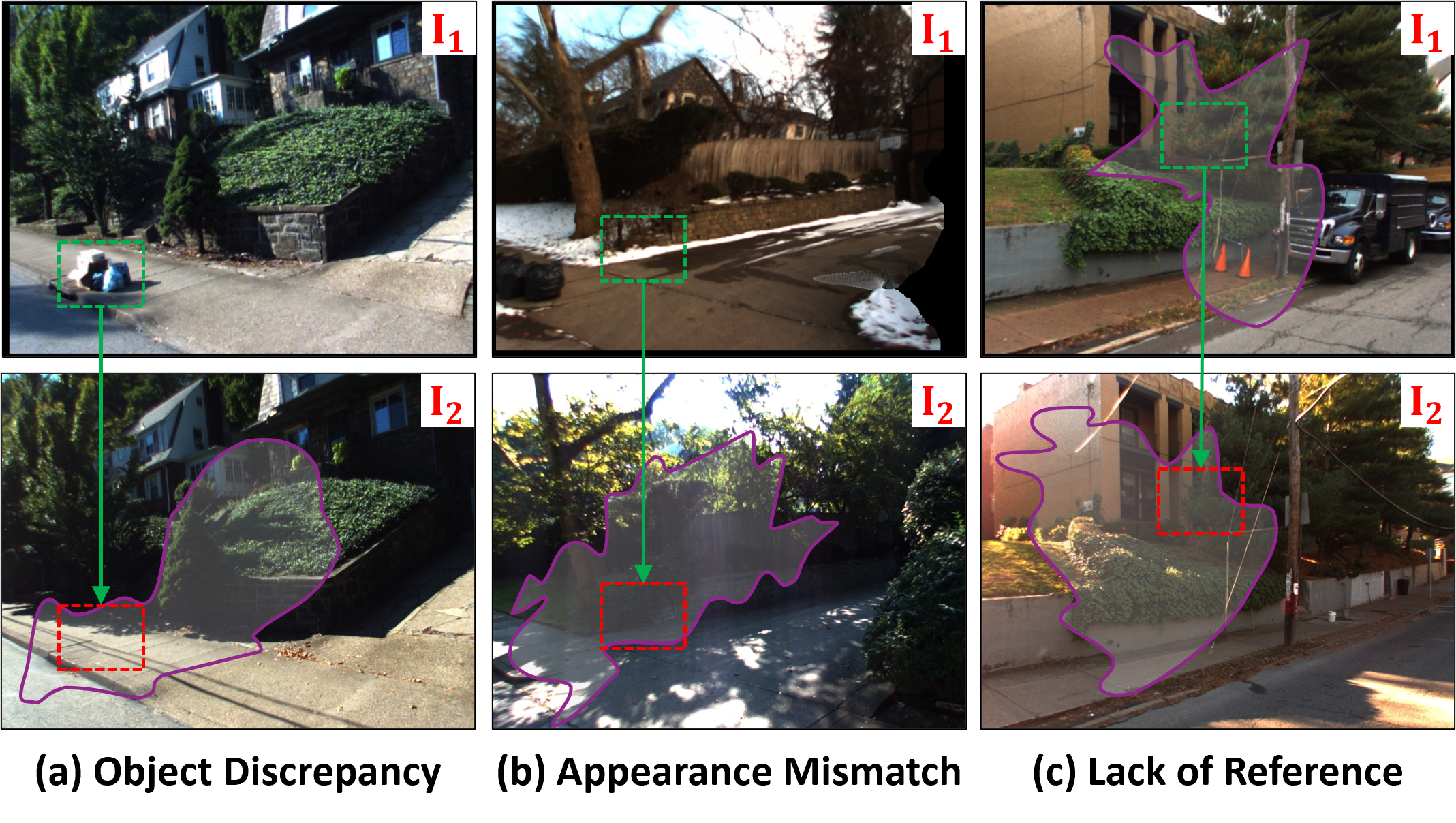}
    \caption{Three main challenges of the proposed time-variant image inpainting tasks. The regions enclosed by \textcolor{Plum}{purple lines} refer to the pixel missing damages.}
    \label{fig:challenges}
\end{figure}

\subsection{Problem Statement}
\label{subsec:formulate}

We consider two images that captured the same scene but with a significant time gap in between (\ie, time-variant images) where object discrepancy and appearance mismatch exist due to the temporal changes.
Different from conventional reference-guided image inpainting, the reference image under the time-variant setup can either be intact or damaged.
The goals are two folds as exemplified in \figref{fig:existing_test}: \ding{182} recovering the damaged target image by referring to the reference image when it is intact, \ie, time-variant reference-guided image inpainting (tvRefInpaint); \ding{183} recovering both images by taking each other as reference when the reference image is also damaged, \ie, time-variant duo-image inpainting (tvDuoInpaint).
Formally, we can formulate both situations with the same equation as
\vspace{-10pt}
\begin{align*}
    &(\hat{\mathbf{I}}_1,\hat{\mathbf{I}}_2) = \phi(\mathbf{I}_1,\mathbf{I}_2) \\
    &\begin{aligned}
        ~\text{s.t.}~~&\mathbf{I}_1 =\text{Shoot}(\mathbf{S}, \mathbf{M}_1, t_1), \\
        &\mathbf{I}_2 =\text{Shoot}(\mathbf{S},\mathbf{M}_2, t_2), \\
        &t_2 - t_1 \gg 0, \\
    \end{aligned}
    \addtocounter{equation}{1}\tag{\theequation}
\end{align*}
where $\mathbf{I}_i=\text{Shoot}(\mathbf{S}, \mathbf{M}_i, t_i)$ means capturing an image of scene $\mathbf{S}$ at time stamp $t_i$ that potentially damaged by pixel missing $\mathbf{M}_i$.
We use $t_2 - t_1 \gg 0$ to indicate the significant time gap between the two images. 
$(\hat{\mathbf{I}}_1, \hat{\mathbf{I}}_2)$ is the reconstructed images and $\phi(\cdot)$ is the desired function to achieve the goal.

\textbf{Challenges.} Compared to the conventional reference-guided image inpainting, the task defined above raises several non-trivial challenges. 
As shown in \figref{fig:challenges} (a) and (b), the main challenges arose, when the reference image is intact, are the object discrepancy and appearance mismatch. 
%
%
%
%
\ding{182} \textbf{Object Discrepancy.} There are clear variations in objects between the time-variant images as the scene changes during time.
For instance, as depicted in \figref{fig:challenges} (a), the object enclosed by the green rectangle in $\mathbf{I}_1$ vanishes in $\mathbf{I}_2$ which indicated by the red rectangle. 
Consequently, what is initially an unlost object in $\mathbf{I}_1$ transforms into interference rather than serving as complementary cues when we attempt to reconstruct $\mathbf{I}_2$ based on $\mathbf{I}_1$.
%
%
\ding{183} \textbf{Appearance Mismatch.} The appearance of the same object in the time-variant images can also be distinct due to day/night variations, weather changes and seasonal shifts, \etc.
As it can be seen from \figref{fig:challenges} (b), the same road corner scene have totally different appearance because of the environmental variation.
Even if there is no obvious object distinction, it is still hard to utilize the complementary pixels in $\mathbf{I}_1$ to recover $\mathbf{I}_2$.
\ding{184} \textbf{Lack of Reference.} Moreover, it becomes even worse when the reference image also suffers from damages.
As shown in \figref{fig:challenges} (c), there is no reference at all for the damaged regions where the content are missing for both target and reference images.
Such a fact further undermines the complementary cues and poses severe challenge.

\begin{figure*}[ht]
    \centering
    \includegraphics[width=\linewidth]{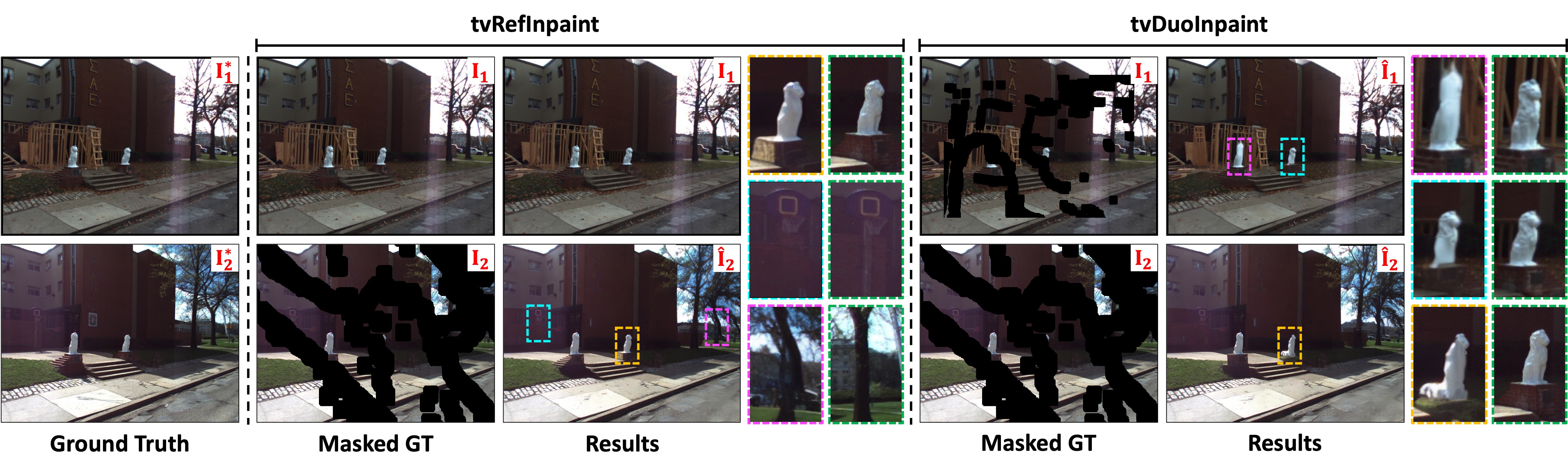}
    \caption{LeftRefill results of exemplar time-variant images under tvRefInpaint and tuDuoInpaint setups. \textcolor{mygreen}{Green} rectangles indicate the regional ground-truth.}
    \label{fig:intuitivexp}
\end{figure*}

\subsection{Empirical Analysis}
\label{subsec:motivation}

Despite the challenges highlighted above, existing reference-guided image inpainting methods can still be intuitively applied to inpaint time-variant images.
%
%
Thus we select LeftRefill~\cite{cao2024leftrefill} as the representative method, which adopts the powerful text-to-image (T2I) model and sets the up-to-date baseline for reference-guided image inpainting, for a case study.
Further experimental details can be found in supplemental material.


\ding{182} \textbf{Time-variant Image Inpainting.}
We first apply LeftRefill to inpaint the time-variant images under conventional reference-guided image inpainting setup.
As the ``tvRefInpaint'' visualized in \figref{fig:intuitivexp}, LeftRefill shows its generative power by fill the damaged regions with smooth contents.
However, we notice that some regions still cannot be restored to the original appearance, \eg, the \textcolor{mycyan}{cyan} rectangle.
%
%
Moreover, LeftRefill also fails to recover the original objects even there are clear references for inpainting, \ie, \textcolor{myyellow}{yellow} and \textcolor{mypink}{pink} regions.
%
%
Next, we apply LeftRefill to inpaint the time-variant images when the reference image is also damaged. 
As the ``tvDuoInpaint'' visualized in \figref{fig:intuitivexp}, it is clear that LeftRefill fails to refill the time-variant images with original content.
Noticeably, as the \textcolor{mypink}{pink} rectangle in the ``Results'' indicated, the lion statue has a clean reference in the counterpart image, but LeftReill still fails to recover its original appearance.
It is even worse when there is only partial or no reference, as exemplified with the \textcolor{myyellow}{yellow} and \textcolor{mycyan}{cyan} rectangles, where LeftRefill totally failed to recover the structure for the lion statue.
In general, we can conclude that the poor performance is mainly caused by the failure of utilizing the counterpart contents as reference.

\begin{figure}[t]
    \centering
    \includegraphics[width=\linewidth]{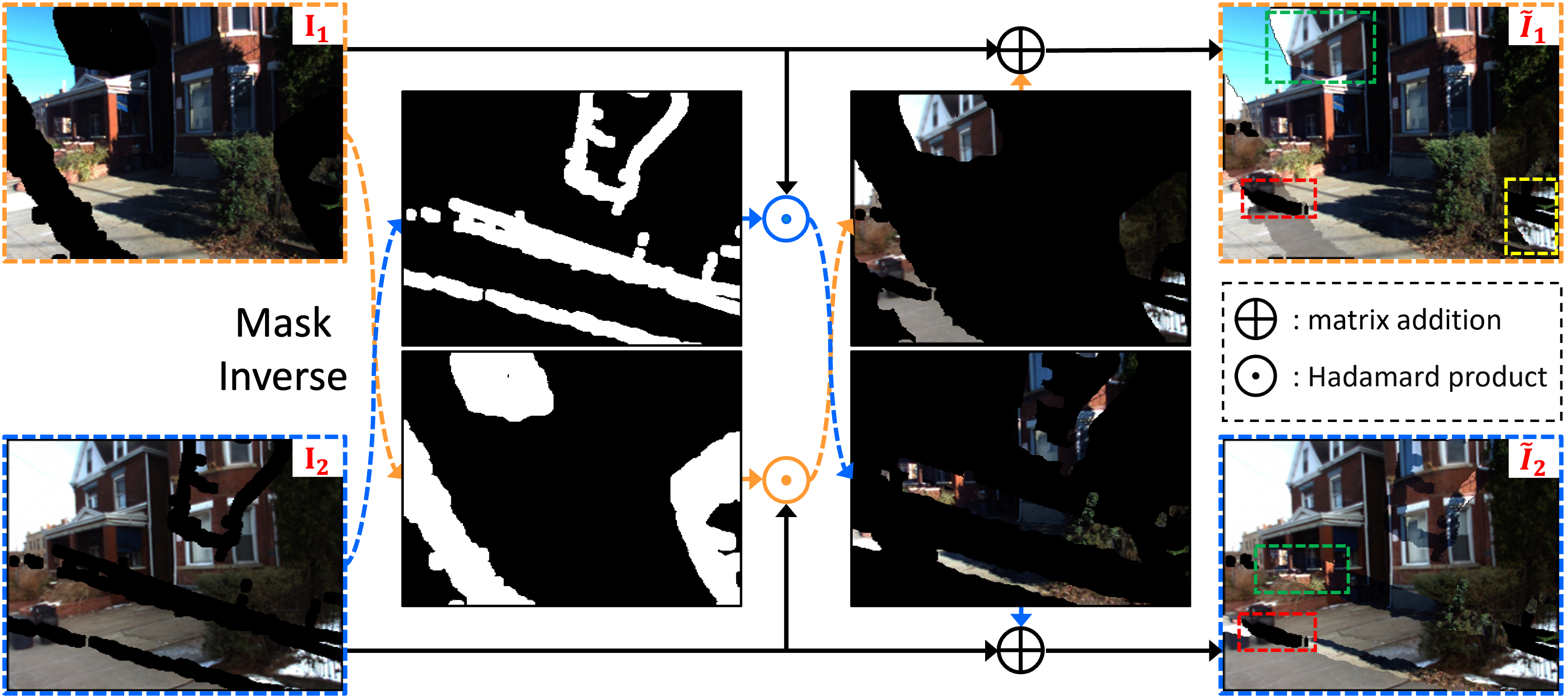}
    \caption{Visualization of naive complementation for the time-variant images. The workflow of the two images are highlighted in \textcolor{myyellow}{yellow} and \textcolor{blue}{blue} dashed line respectively. Note that the masks are inversed for a better view.}
    \label{fig:naive_comp}
\end{figure}
\begin{figure}[t]
    \centering
    \includegraphics[width=\linewidth]{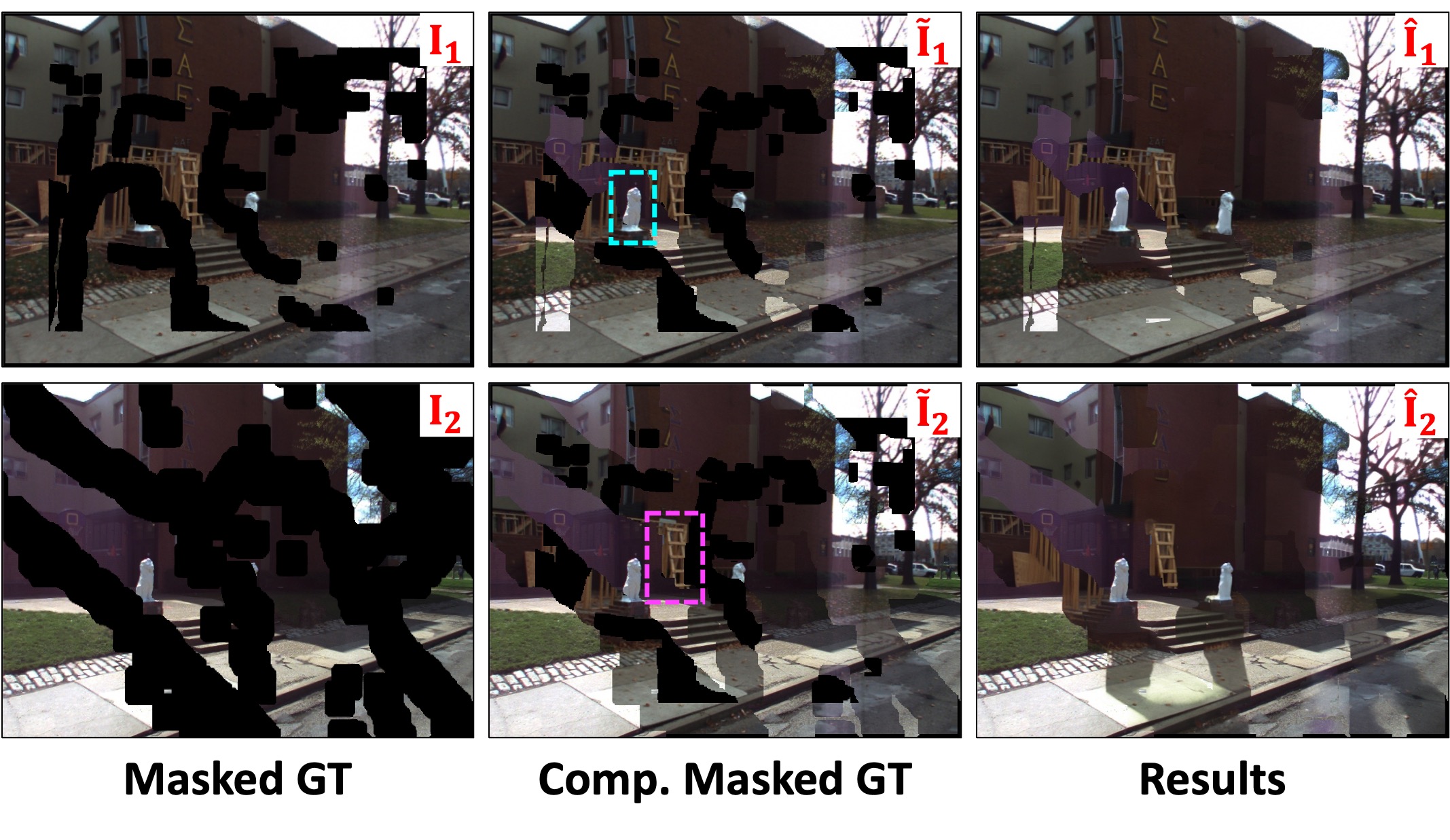}
    \caption{LeftRefill inpainting results with naively complemented time-variant image pair under the tvDuoInpaint setup.}
    \label{fig:compexp}
\end{figure}

\ding{183} \textbf{Naive Complementation.}
Based on the above results, we further conduct experiments by explicitly complementing the images with each other following a naive procedure as 
\begin{equation}
\label{eq:comp}
    \tilde{\mathbf{I}}_1 = \mathbf{I}_1 + {\mathbf{I}}_2 * \bar{\mathbf{M}}_1,~~\tilde{\mathbf{I}}_2 = \mathbf{I}_2 + {\mathbf{I}}_1 * \bar{\mathbf{M}}_2
\end{equation}
where $\bar{\mathbf{M}}_i$ means the inverse of the binary mask, $\tilde{\mathbf{I}}_i$ are the resulting complemented images.
The visualization of such naive complementation process is visualized in \figref{fig:naive_comp}.

By explicitly providing content reference, we expect such naive image complementation can mitigate the problem of inappropriate reference.
Fortunately, as the tuDuoInpaint results shown in \figref{fig:compexp}, the lion statue is restored through naive image complementation (\ie, \textcolor{mycyan}{cyan} rectangle) and the structure is successfully preserved in the final ``Results''.
%
%
Nevertheless, it also can be observed that the significant border artifacts caused by naive complementation are also preserved.
This is due to the object discrepancy introduced by naive image complementation which greatly disturbed the inpainting process.
Take the \textcolor{mypink}{pink} rectangle as an example, the woody building frame should not exist in the final output which is definitely a hindrance for desirable inpainting.
In summary, we see image complementation is of critical importance for time-variant image inpainting, while naively copying and pasting the contents between the images will introduce substantial artifacts which significantly impede the performance. 
%
%
As a result, we absorb our attention in the next to the realization of appropriate image complementation strategy.

\begin{figure*}[t]
    \centering
    \includegraphics[width=\textwidth]{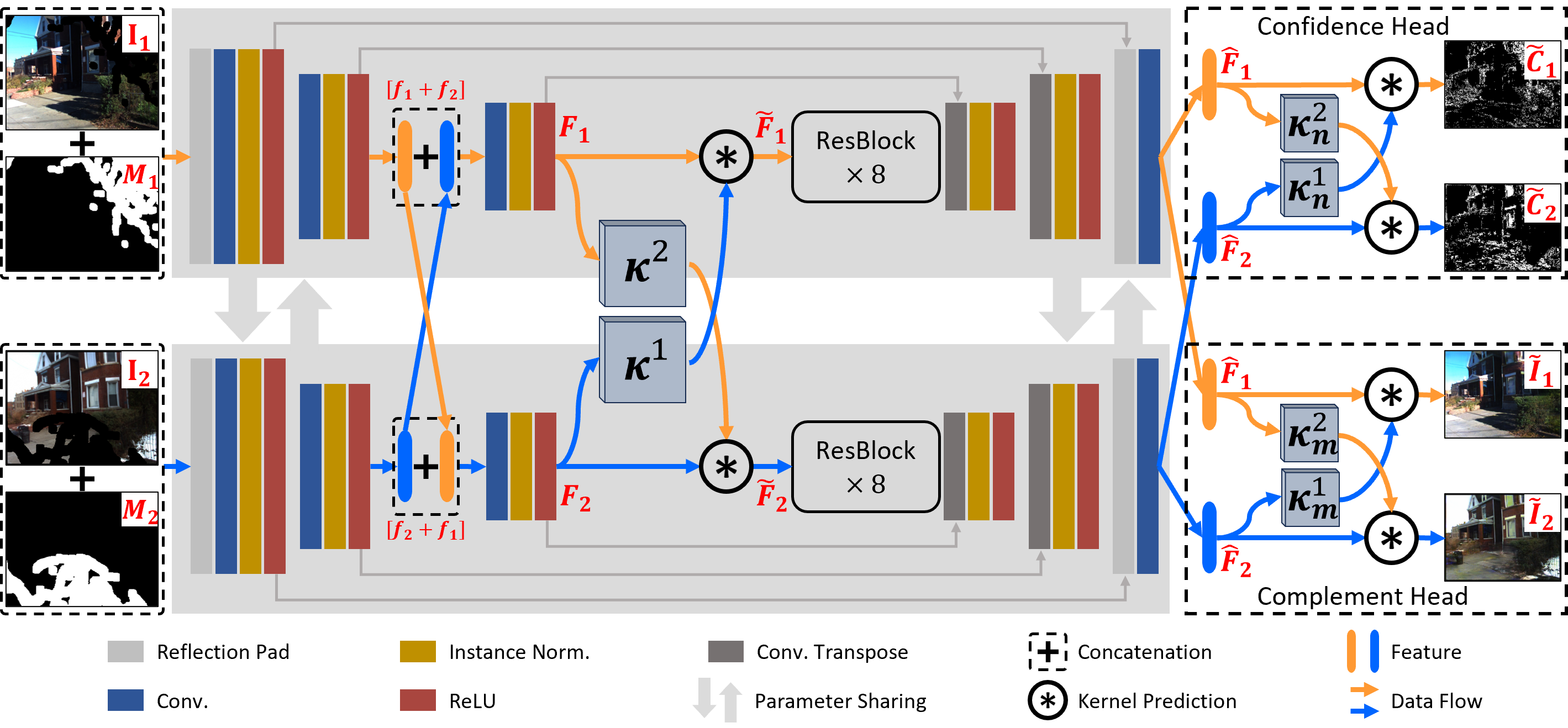}
    \caption{Technical pipelne of the proposed interactive distribution transition estimation module. To make the process be clear, the data flow of the two images are highlighted with \textcolor{BurntOrange}{orange} and \textcolor{blue}{blue} arrow lines respectively. The color of mask $\mathbf{M}_i$ and confidence mask $\mathbf{C}_i$ are inversed for better visualization.}
    \label{fig:idte}
\end{figure*}

\section{Methodology}
\label{sec:method}


Observing the naively complemented images $\mathbf{\tilde{I}}_1$, $\mathbf{\tilde{I}}_2$ in \figref{fig:compexp}, we note that \emph{the empty region and semantic contradiction of the complemented images are the primary deficiencies}.
While the empty region can be filled by generative model (\eg, $\mathbf{\hat{I}}_i$ in \figref{fig:compexp}), the semantic contradiction requires dedicated solution for which we design a module dubbed Interactive Distribution Transition Estimation (InDiTE) to learn how to complement the images with only semantic consistent contents.

%


\subsection{Interactive Distribution Transition Estimation}
\label{subsec:idte}
%
Essentially, what we aim to complement is the content with matching semantics while those semantically contradicted regions should be suppressed to avoid interference.
Although the time-variant images capture the same scene with great temporal change, we note their overall geometric structure remains highly similar
Therefore, we consider the two images as samples from different but geometrically related data distributions, forming a distribution transition process.
Then we explicitly build a module to learn such distribution transition during which the semantically consistent contents are estimated.
As illustrated in \figref{fig:idte}, our designed module consists of a backbone and two specialized heads that detailed as follows.


\subsubsection{Backbone}
Inspired by the success of Siamese network \cite{guo2017learning} in processing paired images, we designed a parameter-shared network with U-Net \cite{unet} skip connection as the image complementing backbone.
With the corrupted time-variant images ${\mathbf{I}}_i$ and corresponding masks $\mathbf{M}_i$ fed as input, our model first encode and merge time-variant images through latent feature concatenation as $[f_i + f_j]$ indicated in \figref{fig:idte}.
By further encoding the merged features, both semantic consistent and contradicted contents are mixed together which can be regarded as feature-level naive complementation.
Drawing inspiration from the remarkable semantic capturing ability of the semantic predictive filtering (SPF) technique \cite{li2022misf}, we then apply it to filter out inconsistent contents by assessing the semantic consistency of the merged features, which can be formulated as
\begin{equation}
    \mathbf{\tilde{F}}_i = \sum_{\mathbf{q}\in \mathcal{N}_\mathbf{p}} \mathbf{K}^j_{\mathbf{p}} [\mathbf{q}-\mathbf{p}]\mathbf{F}_i[\mathbf{q}], ~~~~~i,j\in \{1,2\}, i\neq j,
\label{eq:spf}
\end{equation}
where $\mathbf{p}, \mathbf{q}$ are the coordinates of feature elements and the set $\mathcal{N}_\mathbf{p}$ contains $N^2$ neighboring pixels of $\mathbf{p}$. $\mathbf{\tilde{F}}_i$ is the filtering outputs and $\mathbf{K}^j_\mathbf{p}$ is the kernel to filter the $\mathbf{p}$-th element of $\mathbf{F}_i$ via the neighboring elements, \ie, $\mathcal{N}_\mathbf{p}$. 
By interactively apply the semantic predictive filter, we estimate out the semantically consistent features between the time-variant images.
Then, we feed the resulting features $\mathbf{\tilde{F}}_i$ for decoding whose outputs $\mathbf{\hat{F}}_i$ are further processed by following specilized heads.


\subsubsection{Confidence \& Complement Heads}
As the backbone realized the feature-level time-variant image complementation, we further setup a complement head to get the final image results.
In specific, we apply the same semantic predictive filter to enhance the consistency on image level as
\begin{equation}
    \mathbf{\tilde{I}}_i = \sum_{\mathbf{q}\in \mathcal{N}_\mathbf{p}} \mathbf{K}^j_{m,\mathbf{p}} [\mathbf{q}-\mathbf{p}]\mathbf{\hat{F}}_i[\mathbf{q}],
\end{equation}
where $\mathbf{\tilde{I}}_i$ is the final complemented images and $\mathbf{\hat{F}}_i$ is the output feature from backbone.
Now the semantic consistent complementation is finalized while the lack of reference challenge is still under-resolved.
As discussed in \secref{subsec:motivation}, existing generative models are fully capable of fill the empty regions that caused by lack of reference with plausible contents.
However, naively adopt the mask intersection $\mathbf{M}_1\odot\mathbf{M}_2$ for indication will erase the complementation efforts from backbone, thus we build the confidence head to learn a fine-grained confidence mask for indicating the regions that require further generation.
Similar to the complement head, we keep using semantic predictive filter for the confidence head as
\begin{equation}
    \mathbf{\tilde{C}}_i = \sum_{\mathbf{q}\in \mathcal{N}_\mathbf{p}} \mathbf{K}^j_{n,\mathbf{p}} [\mathbf{q}-\mathbf{p}]\mathbf{\hat{F}}_i[\mathbf{q}].
\end{equation}
where $\mathbf{\tilde{C}}_i$ is the resulting confidence map. Then we get the pixel-wise confidence mask by binarizing the $\mathbf{\tilde{C}}_i$ with 
\begin{equation}
    \mathbf{C}_i = \left\{ 
            \begin{array}{rl} 
                1, & \mathbf{\tilde{C}}_i>\tau \\ 
                0, & \text{otherwise} \\
            \end{array}\right.
\end{equation}
where $\tau$ is the threshold for binarizing.
%
Intuitively, the mask $\mathbf{C}_i$ is expected to inform the subsequent generative model which part of the complemented contents should be reserved and which needs further processing.
In fact, as we will see in the experiments, such a combination of head design generally ensures further suppression of potential semantic inconsistency thus reducing its negative impact on the final inpainted results.


%
\begin{figure*}[t]
    \centering
    \includegraphics[width=\linewidth]{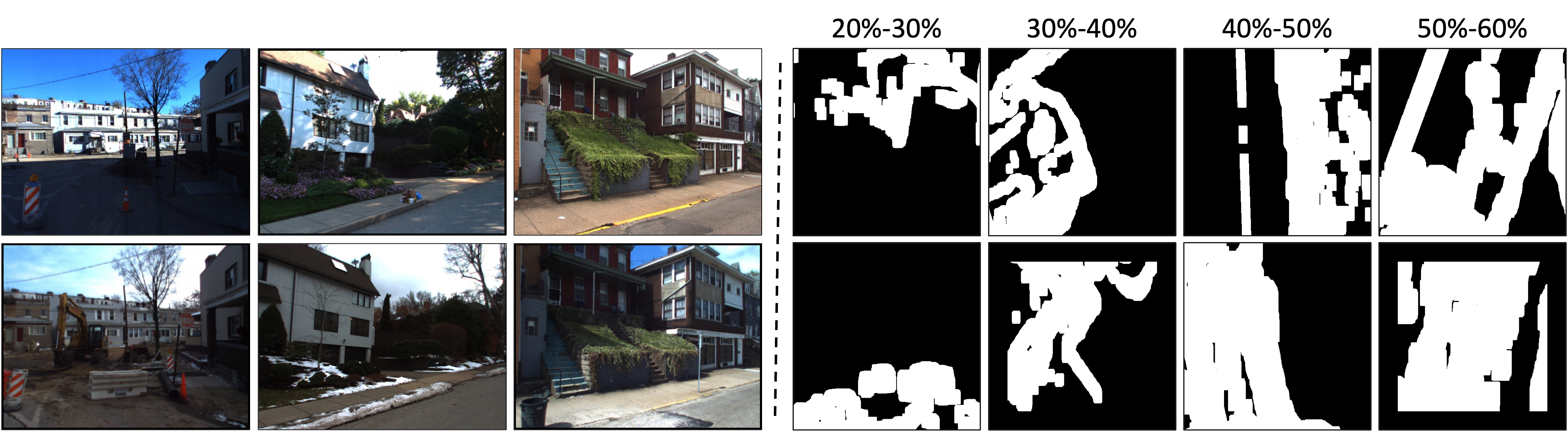}
    \caption{Illustration of representative time-variant image pairs and mask pairs at varying levels of degradation, adopted to construct our TAMP-Street dataset.}
    \label{fig:dataset}
\end{figure*}

\subsection{Interactive Distribution Transition-driven Diffusion}
\label{subsec:idteapply}


With the complemented images and the corresponding confidence mask, now we can proceed to apply generative model for addressing the lack of reference issue.
In specific, we build on DDNM \cite{wang2022ddnm} and propose the Interactive Distribution Transition Estimation-driven Diffusion (InDiTE-Diff) for generation.
%
%
%
Following the convention of diffusion notations, we use $\mathbf{x}^i_0$ and $\mathbf{x}^i_T$ to denote the final inpainted results $\mathbf{\hat{I}_i}$ and the inputs (Hadamard product $\mathbf{\tilde{I}}_i \odot \mathbf{C}_i$ in our case) respectively. Then we setup the sampling process as 
\begin{align}
\begin{split}
    \mathbf{x}^i_{t-1} =& \frac{\sqrt{\bar{\alpha}_{t-1}}\beta_t}{1-\bar{\alpha}_t} \hat{\mathbf{x}}^i_{0|t} + \frac{\sqrt{\alpha_t}(1-\bar{\alpha}_{t-1})}{1-\bar{\alpha}_t} \mathbf{x}^i_t \\&+ \sigma_t \mathbf{\epsilon}, ~~~\mathbf{\epsilon}\sim \mathcal{N}(0,\mathbf{1}),
\label{eq:diff}
\end{split}
\end{align}
where $\beta_t$ is the predefined scale factor and $\alpha_t=1-\beta_t$, $\sigma_t^2=\frac{1-\bar{\alpha}_{t-1}}{1-\bar{\alpha_t}}\beta_t$. 
Following DDNM, we have the modified diffusion sampling $\mathbf{\hat{x}}^i_{0|t} = \mathbf{A}^\dagger(\mathbf{\tilde{I}}_i\odot\mathbf{C}_i)+(\mathbf{1}-\mathbf{A}^\dagger\mathbf{A})\mathbf{x}^i_{0|t}$ where the $\mathbf{x}^i_{0|t}$ denotes the estimated $\mathbf{x}^i_0$ at time-step $t$ and $\mathbf{A}$, $\mathbf{A}^\dagger$ is the degradation operator and corresponding sudo-inverse respectively.
Moreover, as content authenticity is of critical importance for image inpainting task, we further add cross-reference between the time-variant images during diffusion sampling.
In specific, we follow \cite{dda} and interactively apply low-pass filter $\phi_D(\cdot)$ to the sampling process
\begin{equation}
    \mathbf{\hat{x}}^i_{t-1} \leftarrow \mathbf{x}^i_{t-1} - \mathbf{\omega}\bigtriangledown_{\mathbf{x}^i_t} ||\phi_D(\mathbf{\tilde{I}}_j\odot\mathbf{C}_j)-\phi_D(\hat{\mathbf{x}}^i_{0|t}\odot\mathbf{C}_j)||_2, 
\label{eq:interact}
\end{equation}
which essentially requires the generation of the duo-image to align with each other on low frequency for every time step.
%

\subsection{Implementation Details}
\subsubsection{Loss Function}
To get high-fidelity and semantic consistent results, we follow \cite{nazeri2019edgeconnect} to train the networks with four loss functions, \ie, L1 loss, GAN loss, style loss, and perceptual loss. 
In specific, given the complemented results $\mathbf{\tilde{I}_i}$ and the ground truth $\mathbf{I}_i^*$ of time-variant images, we have the loss function for complement head as
\begin{equation}
    \mathcal{L}(\mathbf{\tilde{I}_i, \mathbf{I}_i^*})=\lambda_1\mathcal{L}_1^m + \lambda_2\mathcal{L}_{gan}^m + \lambda_3\mathcal{L}_{pct}^m + \lambda_4\mathcal{L}_{sty}^m.
\label{eq:com_head}
\end{equation}
Similarly, by setting the ground truth as $\mathbf{\tilde{C}}_i^* =\mathbf{I_i^*-\mathbf{\tilde{I}}}_i$, we calculate the loss for the predicted confidence map $\mathbf{\tilde{C}}_i$ as
\begin{equation}
    \mathcal{L}(\mathbf{\tilde{C}}_i, \mathbf{\tilde{C}}_i^*)=\lambda_1\mathcal{L}_1^n + \lambda_2\mathcal{L}_{gan}^n + \lambda_3\mathcal{L}_{pct}^n + \lambda_4\mathcal{L}_{style}^n.
\label{eq:cof_head}
\end{equation}
The $\lambda_i$ are the weighting parameters for balancing the losses during the training of both heads.
Please find the detail definitions of the loss functions in \cite{nazeri2019edgeconnect}.

\subsubsection{Diffusion Sampling}
Following DDNM \cite{wang2022ddnm}, we set the degradation operator as $\mathbf{A}=\mathbf{x}_{0|t}^i\odot\mathbf{M}_i$. As for the low-pass filter $\phi_D(\cdot)$, we follow \cite{choi2021ilvr} \cite{dda} to apply a sequence of downsampling and upsampling operations with the scale factor $D$. Please refer to \cite{dda} for the detail definition.


\begin{table}
    \vspace{10pt}
    \caption{Comparison of challenges involved in TAMP-Street and the datasets from existing reference-guided image inpainting studies.}
    \label{tab:datadif}
    \resizebox{0.5\textwidth}{!}{%
    \begin{tabular}{c|cccc}
        \toprule
        \diagbox{Data Source}{Challenge} & \makecell[c]{\makecell[c]{Geometric\\Misalignment}} & \makecell[c]{Object\\Discrepancy} & \makecell[c]{Appearance\\Mismatch} & \makecell[c]{Reference\\Lack} \cr
        \midrule
        TransFill\cite{zhou2021transfill}       & \cmark & \xmark & \xmark & \xmark \cr
        \rule{0pt}{3ex}TransRef\cite{liao2023transref}        & \cmark & \xmark & \cmark & \xmark \cr
        \rule{0pt}{3ex}LeftRefill\cite{cao2024leftrefill}      & \cmark & \cmark & \xmark & \xmark \cr
        \rule{0pt}{3ex}TAMP-Stree (ours)     & \cmark & \cmark & \cmark & \cmark \cr
        \bottomrule
    \end{tabular}
    }
\end{table}

\begin{table*}[ht]
    \caption{The quantitative results of the baseline methods and our proposed InDiTE-Diff for inpainting time-variant images under conventional reference-guided image inpainting setup, \ie, tvRefInpaint, on TAMP-Street dataset. The overall best results are highlighted in \textcolor{red}{red} font, while the second best in \textbf{bold} font.}
    \label{tab:ref_result}
    \centering
    \setlength{\tabcolsep}{6pt}
    \resizebox{0.75\linewidth}{!}{
    \begin{tabular}{c|cccc|cccc}
        \toprule
        \multicolumn{1}{c|}{\diagbox{Method}{Mask}} & 20\%-30\% & 30\%-40\% & 40\%-50\% & 50\%-60\% & 20\%-30\% & 30\%-40\% & 40\%-50\% & 50\%-60\% \cr
        \bottomrule
        \toprule
            \multicolumn{1}{c|}{} & \multicolumn{4}{c|}{PSNR $\uparrow$} & \multicolumn{4}{c}{SSIM $\uparrow$} \cr
        \cmidrule{1-9}
        \textit{reference} & \multicolumn{4}{c|}{\textit{44.8920}} & \multicolumn{4}{c}{\textit{0.9963}} \cr
        \cmidrule{1-9}
        TransFill     & 26.7725 & 28.8931 & 25.9461 & 25.2256 & 0.8771 & 0.8800 & 0.8441 & 0.8459 \cr
        TransRef      & 28.6931 & 28.8201 & 26.7362 & 24.9292 & 0.8810 & 0.8889 & \textbf{0.8772} & 0.8510 \cr
        LeftRefill    & \textcolor{red}{30.9928} & \textcolor{red}{30.3511} & \textbf{27.8821} & \textbf{25.7842} & \textcolor{red}{0.9438} & \textcolor{red}{0.9229} & 0.8236 & \textbf{0.8539} \cr
        InDiTE-Diff (ours)        & \textbf{30.9782} & \textbf{29.9718} & \textcolor{red}{28.8900} & \textcolor{red}{26.7274} & \textbf{0.9338} & \textbf{0.9071} & \textcolor{red}{0.89406} & \textcolor{red}{0.8810} \cr
        \midrule
            \multicolumn{1}{c|}{} & \multicolumn{4}{c|}{$L_1$ $\downarrow$} & \multicolumn{4}{c}{LPIPS $\downarrow$} \cr
        \cmidrule{1-9}
        \textit{reference} & \multicolumn{4}{c|}{\textit{0.0007}} & \multicolumn{4}{c}{\textit{0.0100}} \cr
        \cmidrule{1-9}
        TransFill     & 0.0292 & 0.0262 & 0.0351 & 0.0380 & 0.0233 & 0.0182 & 0.0277 & 0.0295 \cr
        TransRef      & 0.0253 & 0.0282 & 0.0322 & \textbf{0.0374} & 0.0179 & 0.0176 & 0.0277 & \textbf{0.0293} \cr
        LeftRefill    & \textcolor{red}{0.0155} & \textcolor{red}{0.0195} & \textbf{0.0290} & 0.0382 & \textcolor{red}{0.0142} & \textcolor{red}{0.0150} & \textbf{0.0186} & 0.0315 \cr
        InDiTE-Diff (ours) & \textbf{0.0181} & \textbf{0.0205} & \textcolor{red}{0.0289} & \textcolor{red}{0.0296} & \textbf{0.0155} & \textbf{0.0175} & \textcolor{red}{0.0172} & \textcolor{red}{0.0181} \cr
        \bottomrule
    \end{tabular}
    }
\end{table*}
\begin{table*}[ht]
    \caption{The quantitative results of the baseline methods and our proposed InDiTE-Diff for solving the time-variant duo-image inpainting task, \ie, tvDuoInpaint, on TAMP-Street dataset. The overall best results are highlighted in \textcolor{red}{red} font, while the second best in \textbf{bold} font.}
    \label{tab:tamp_result}
    \centering
    \resizebox{\linewidth}{!}{
    \begin{tabular}{c|cccc|cccc}
        \toprule
        \multicolumn{1}{c|}{\diagbox{Method}{Mask}} & 20\%-30\% & 30\%-40\% & 40\%-50\% & 50\%-60\% & 20\%-30\% & 30\%-40\% & 40\%-50\% & 50\%-60\% \cr
        \bottomrule
        \toprule
            \multicolumn{1}{c|}{} & \multicolumn{4}{c|}{PSNR $\uparrow$} & \multicolumn{4}{c}{SSIM $\uparrow$} \cr
        \cmidrule{1-9}
        TransFill       & 19.3377/20.1026 & 17.2356/17.6748 & 15.1677/15.8720 & 13.1019/13.7011 & 0.7800/0.7721 & 0.7233/0.7319 & 0.6807/0.6925 & 0.5567/0.5583 \cr
        TransRef        & 18.9164/22.7751 & 17.2486/20.9556 & 15.9129/19.2969 & 14.6797/\textbf{16.2012} & 0.8322/0.8617 & 0.7623/0.8067 & 0.6896/0.7438 & 0.6157/0.6298 \cr
        LeftRefill      & \textbf{23.1727/23.1517} & \textbf{21.6623/21.9740} & \textbf{19.7019/19.5421} & \textbf{15.2239}/15.9228 & \textbf{0.8949}/\textcolor{red}{0.9054} & \textcolor{red}{0.8613}/\textbf{0.8487} & \textbf{0.7950}/\textcolor{red}{0.8123} & \textbf{0.6962/0.6971} \cr
        InDiTE-Diff (ours) & \textcolor{red}{26.2677/26.5959} & \textcolor{red}{24.5230/24.7559} & \textcolor{red}{22.7704/23.3243} & \textcolor{red}{20.4491/20.9545} & \textcolor{red}{0.8977}/\textbf{0.8959} & \textbf{0.8533}/\textcolor{red}{0.8513} & \textcolor{red}{0.8060}/\textbf{0.8040} & \textcolor{red}{0.7278/0.7210} \cr
        \midrule
            \multicolumn{1}{c|}{} & \multicolumn{4}{c|}{$L_1$ $\downarrow$} & \multicolumn{4}{c}{LPIPS $\downarrow$} \cr
        \cmidrule{1-9}
        TransFill       & 0.1840/0.1814 & 0.1930/0.1951 & 0.3031/0.2894 & 0.3877/0.3503 & 0.1126/0.1184 & 0.1587/0.1639 & 0.2045/0.2121 & 0.2669/0.2767 \cr
        TransRef        & \textbf{0.0819/0.0458} & \textbf{0.1159/0.0658} & \textbf{0.1515/0.0897} & \textbf{0.1976/0.1559} & 0.1631/0.1293 & 0.2279/0.1809 & 0.2883/0.2337 & 0.3482/0.3307 \cr
        LeftRefill      & 0.0851/0.0883 & 0.1340/0.1239 & 0.1746/0.1821 & 0.3034/0.2744 & \textbf{0.0283/0.0290} & \textbf{0.0533/0.0520} & \textbf{0.0860/0.0783} & \textbf{0.1002/0.0925} \cr
        InDiTE-Diff (ours) & \textcolor{red}{0.0268/0.0264} & \textcolor{red}{0.0380/0.0379} & \textcolor{red}{0.0517/0.0495} & \textcolor{red}{0.0759/0.0739} & \textcolor{red}{0.0177/0.0182} & \textcolor{red}{0.0284/0.0269} & \textcolor{red}{0.0386/0.0407} & \textcolor{red}{0.0869/0.0736} \cr
        \bottomrule
    \end{tabular}
    }
\end{table*}

\section{TAMP-Street Dataset}
\label{sec:data}

As discussed in \secref{sec:intro}, existing datasets cannot meet the requirements of time-variant image inpainting problem setup.
\tableref{tab:datadif} , 
Therefore, we assemble the TAMP-Street\footnote{\url{https://drive.google.com/drive/folders/1frK37q4CZ2tDUc5_x-65N2wOXEd2oJcL?usp=sharing}} dataset based on existing image and mask source to evaluate inpainting models under TAMP setup.
For a clear comparison, we summarize the main differences between the assembled TAMP-Street and the data utilized by existing researches in .
Further visualization of TAMP-Street dataset can be found in \figref{fig:dataset}.

\subsubsection{Time-variant Images} As the time-variant image inpainting setup detailed in \secref{subsec:formulate}, the images under TAMP setup are required to be taken from the same scene with a large time gap between them. To meet such specifications, we adopt the images from VL-CMU-CD \cite{alcantarilla2018street} as our image basis for dataset building. The VL-CMU-CD dataset was originally proposed for object changing detection which documents one year of diverse urban street transformations. Specifically, there are a total of 1,362 pairs of street view images all of which have a size of 1024$\times$768. Following the original split, 816 pairs of images are utilized for training, 256 pairs of images for validation, and 290 pairs for testing.

\subsubsection{Irregular Mask for TAMP} To simulate the random pixel damage, we build our mask pairs based on the irregular hole mask introduced by \cite{liu2018mask}. Originally, the masks are split by mask ratio from 0\%-60\% with 20\% interval. Benefiting from the advancements of the generative model \cite{zhang2023copaint}, the restoration of small pixel missing has become an effortless task (see supplemental material). Thus we only adopt mask ratio 20\%-60\% for TAMP task and set the interval as 10\%. In detail, we randomly pair the masks under each ratio and totally build 1600 pairs of masks for testing (400 under each mask ratio), 5600 and 800 pairs of masks for training and evaluation respectively. Note that we build our mask pairs by only utilizing the testing mask set of \cite{liu2018mask}, which is enough to cover the image pairs.
Note that in the original mask source (\ie, the test split of \cite{liu2018mask}), the masks are also identified as ``with border constraint'' and ``without border constraint''.
The masks for our TAMP-Street dataset are randomly sampled from the mask source without considering the difference of mask border as both situations can appear in the real world.

\begin{figure*}[ht]
    \centering
    \includegraphics[width=\textwidth]{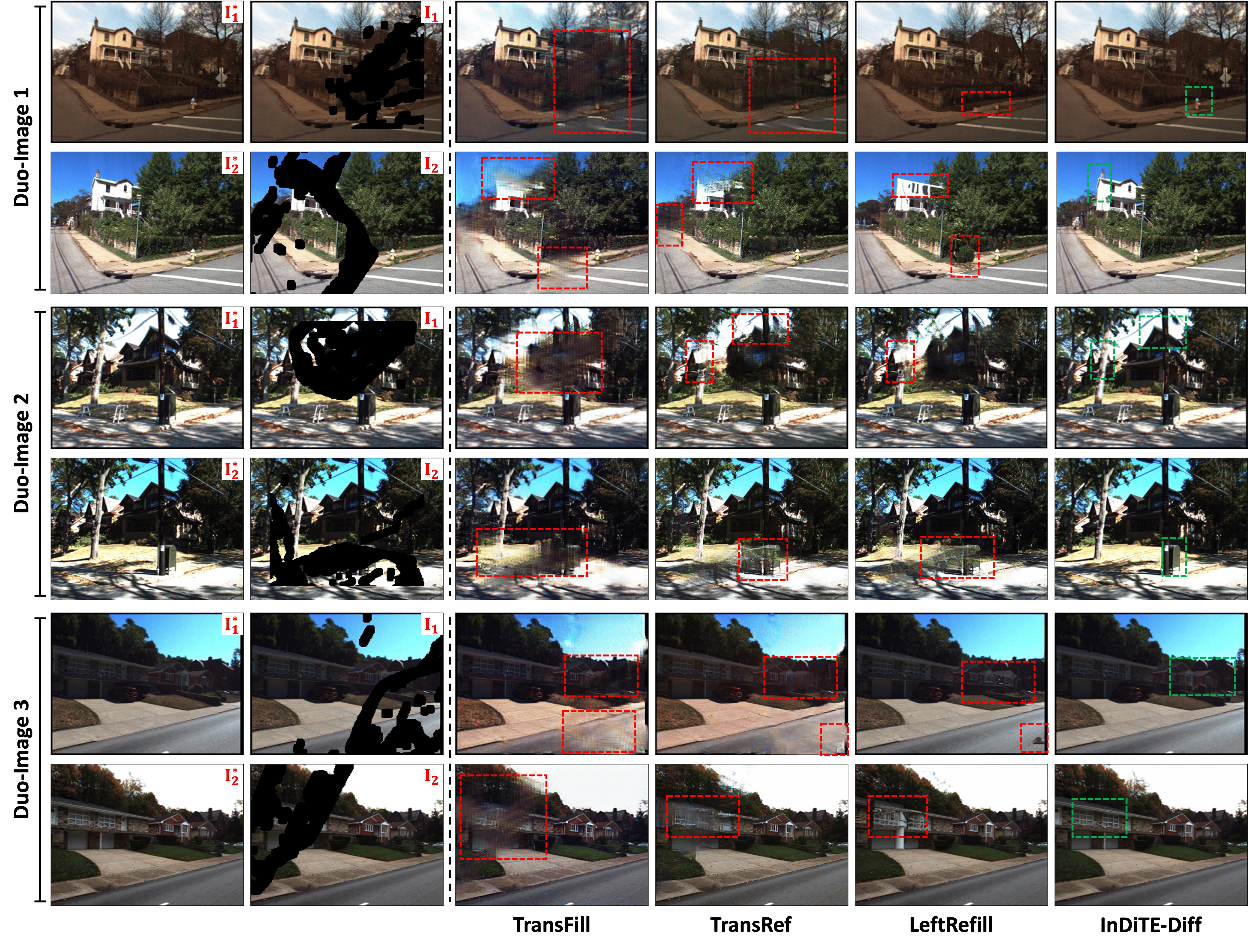}
    \caption{Visualization of existing reference-guided image inpainting methods and our InDiTE-Diff for inpainting typical target time-variant images under tvDuoInpaint setup. The defects are highlighted with \textcolor{red}{red} rectangles, and the \textcolor{green}{green} rectangles indicate corresponding InDiTE results.}
    \label{fig:tamp_vis}
\vspace{-5pt}
\end{figure*}

\section{Experiments}
\label{sec:exp}


\subsection{Experimental Setup}
\label{subsec:setup}

\subsubsection{Environment}
The proposed model is implemented with Python 3.8.18 based on PyTorch 1.7.1 and cuda11.
All the experiments are conducted on the same workstation with an AMD EPYC 7763 64-Core CPU,  504GB RAM, and four NVIDIA A40 GPUs (46GB memory each), where the operating system is Ubuntu 20.04.

\subsubsection{Baselines}
We compare the proposed InDiTE method with three SOTA reference-guided image inpainting methods, \ie, TransFill \cite{zhou2021transfill}, TransRef \cite{liao2023transref} and LeftRefill \cite{cao2024leftrefill}.
In specific, TransFill \cite{zhou2021transfill} represents traditional registration pipeline for image inpainting,
TransRef \cite{liao2023transref} adopts Transformer architecture and belongs to the learning-based image inpainting method,
and LeftRefill \cite{cao2024leftrefill} utilizes powerful text-to-image foundation model as the backbone for reference-guided image inpainting.


\subsubsection{Metrics} Following the convention \cite{li2022misf}, we conduct evaluation with four metrics which are peak signal-to-noise ratio (PSNR), structural similarity index (SSIM), perceptual similarity (LPIPS) and $L_1$. PSNR, SSIM, and $L_1$ measure the quality of the recovered image, and LPIPS measures the perceptual consistency between the recovered image and ground truth. 
For the tvRefInpaint setup, we only report the results for the target image under each mask ratio as the reference images are kept intact under this time-variant setting.
For the tvDuoInpaint, we report all the metrics for both reference and target images in separate for a clear comparison, \eg, $L_1^{\mathbf{I}_1}/L_1^{\mathbf{I}_2}$.

\subsubsection{Model Training}
%
%
%
We train InDiTE on the training split of our TAMP-Street dataset, where we follow \cite{nazeri2019edgeconnect} and train InDiTE with four loss functions as shown in \reqref{eq:com_head} and \reqref{eq:cof_head}.
Following \cite{li2022misf}, we set the weighting hyper-parameters of the four losses as $\lambda_1=1$, $\lambda_2=\lambda_3=0.1$ and $\lambda_4=250$.
The discriminator of GAN utilized for computing the loss is also set to the same as \cite{li2022misf}.
During optimization, we adopt Adam optimizer with $\beta_1=0, \beta_2=0.9$ and a learning rate $1\times 10^{-4}$ to train InDiTE for 200 epoch in total.
We evaluate InDiTE on the evaluation split of the datasets and mask pairs every 5 epochs and save the model with the best PSNR as the resulting model for testing. 
For the dataset, both images and masks are resized into 256$\times$256 where the images are normalized into $[-1,1]$ and the masks are binarized into $\{0,1\}$.
%

To keep the fairness of comparison, all the baselines are retrained on TAMP-Street training data then evaluated with the testing split, during which we strictly follow each baseline's original released code, \ie, TransRef\footnote{\url{https://github.com/Cameltr/TransRef}}, LeftRefill\footnote{\url{https://github.com/ewrfcas/LeftRefill}}.
Expect for TransFill \cite{zhou2021transfill} which can only be evaluated by sending the data to their server for evaluation.
%


\subsection{Time-variant Reference-guided Image Inpainting}
To show that our proposed InDiTE-Diff is competitive to existing methods, we first conduct experiments with the conventional reference-guided image inpainting setup on the TAMP-Street dataset. 
The statistical results are summarized in \tableref{tab:ref_result}.
%
%
\ding{182} Firstly, we note that InDiTE-Diff demonstrates a consistent advantage over the baselines.
Although the improvements are less pronounced when the mask ratio is relatively smaller (\eg, $30\%-40\%$), the performance is improved significantly as the mask ratio increases, \ie, 1.3693\% and 1.0767\% for $40\%-50\%$ and $50\%-60\%$ respectively.
This indicates that under the time-variant setting, our approach—particularly in high-ratio damage scenarios—achieves more effective utilization of the reference image.
%
%
\ding{183} On the other hand, we observe that, compared to results obtained on clean reference images, existing methods under the time-variant setting struggle to achieve comparable levels of restoration for damaged images.
This highlights the inherent challenge of the time-variant scenario.
However, as discussed in \secref{sec:intro}, the time-variant setting is a more realistic and application-relevant image restoration setting.
The current results suggest that more effective solutions are still required to address this problem.
In general, our proposed InDiTE-Diff achieves competitive reference-guided image inpainting for time-variant images.
Next we proceed to experiment under time-variant duo-image inpainting setup where more serious issues are posed to challenge existing inpainting paradigm.



\begin{figure*}[ht]
    \begin{minipage}[t]{.58\linewidth} \vspace{0pt}
    \centering
    \captionof{table}{Existing methods' PSNR results before and after InDiTE boosting on the TAMP-Street dataset. The improved performance are highlighted in \textcolor{red}{red}.}
    \resizebox{\linewidth}{!}{
    \begin{tabular}{c|cccc}
        \toprule
        \multicolumn{1}{c|}{\diagbox{Method}{Mask}} & 20\%-30\% & 30\%-40\% & 40\%-50\% & 50\%-60\% \cr
        \bottomrule
        \toprule
            \multicolumn{1}{c|}{} & \multicolumn{4}{c}{Time-variant Reference-guided Image Inpainting}  \cr
        \cmidrule{1-5}
        TransFill                    & 26.7725/28.7206 & 28.8931/28.9077 & 25.9461/27.6339 & 25.2256/25.4573   \cr
        \textbf{TransFill-InDiTE}      & \textcolor{red}{27.2018/28.9009} & \textcolor{red}{28.9000}/28.7933 & \textcolor{red}{26.2107/27.6900} & \textcolor{red}{25.3539/25.5621} \cr
        TransRef                     & 28.6931/28.7706 & 28.8201/29.3552 & 26.7362/27.0887 & 24.9292/25.6901   \cr
        \textbf{TransRef-InDiTE}       & \textcolor{red}{28.7072}/28.7055 & \textcolor{red}{29.0312}/29.2236 & \textcolor{red}{27.1104/27.3259} & \textcolor{red}{25.0216/25.7132}  \cr
        LeftRefill                   & 30.9928/30.9830 & 30.3511/30.5271 & 27.8821/28.3538 & 25.7842/25.8239  \cr
        \textbf{LeftRefill-InDiTE}     & \textcolor{red}{31.0226/30.9870} & 30.3321/\textcolor{red}{30.6383} & \textcolor{red}{28.0130/28.3541} & 25.7709/\textcolor{red}{26.1319}  \cr
        \midrule
            \multicolumn{1}{c|}{} & \multicolumn{4}{c}{Time-variant Duo-image Inpainting}\cr
        \cmidrule{1-5}
        TransFill                     & 19.3377/20.1026 & 17.2356/17.6748 & 15.1677/15.8720 & 13.1019/13.7011  \cr
        \textbf{TransFill-InDiTE}       & \textcolor{red}{20.7309/21.4531} & \textcolor{red}{18.6612/19.0100} & \textcolor{red}{17.1427/17.6803} & \textcolor{red}{15.1152/15.2268}  \cr
        TransRef                      & 18.9164/22.7751 & 17.2486/20.9556 & 15.9129/19.2969 & 14.6797/16.2012 \cr
        \textbf{TransRef-InDiTE}        & \textcolor{red}{20.5575/23.8705} & \textcolor{red}{18.9886/22.6911} & \textcolor{red}{17.4543/20.5932} & \textcolor{red}{16.8724/18.5080}  \cr
        LeftRefill                    & 23.1727/23.1517 & 21.6623/21.9740 & 19.7019/19.5421 & 15.2239/15.9228  \cr
        \textbf{LeftRefill-InDiTE}      & \textcolor{red}{23.5043/24.9533} & \textcolor{red}{22.8759/23.9154} & \textcolor{red}{20.8503/20.4831} & \textcolor{red}{17.1051/17.9848}  \cr
        \bottomrule
    \end{tabular}
    \label{tab:itdeboost}
    }
    \end{minipage}
    \hfill
    \begin{minipage}[t]{0.4\linewidth} \vspace{0pt}
        \centering
        \vspace{-5pt}
        \includegraphics[width=\linewidth]{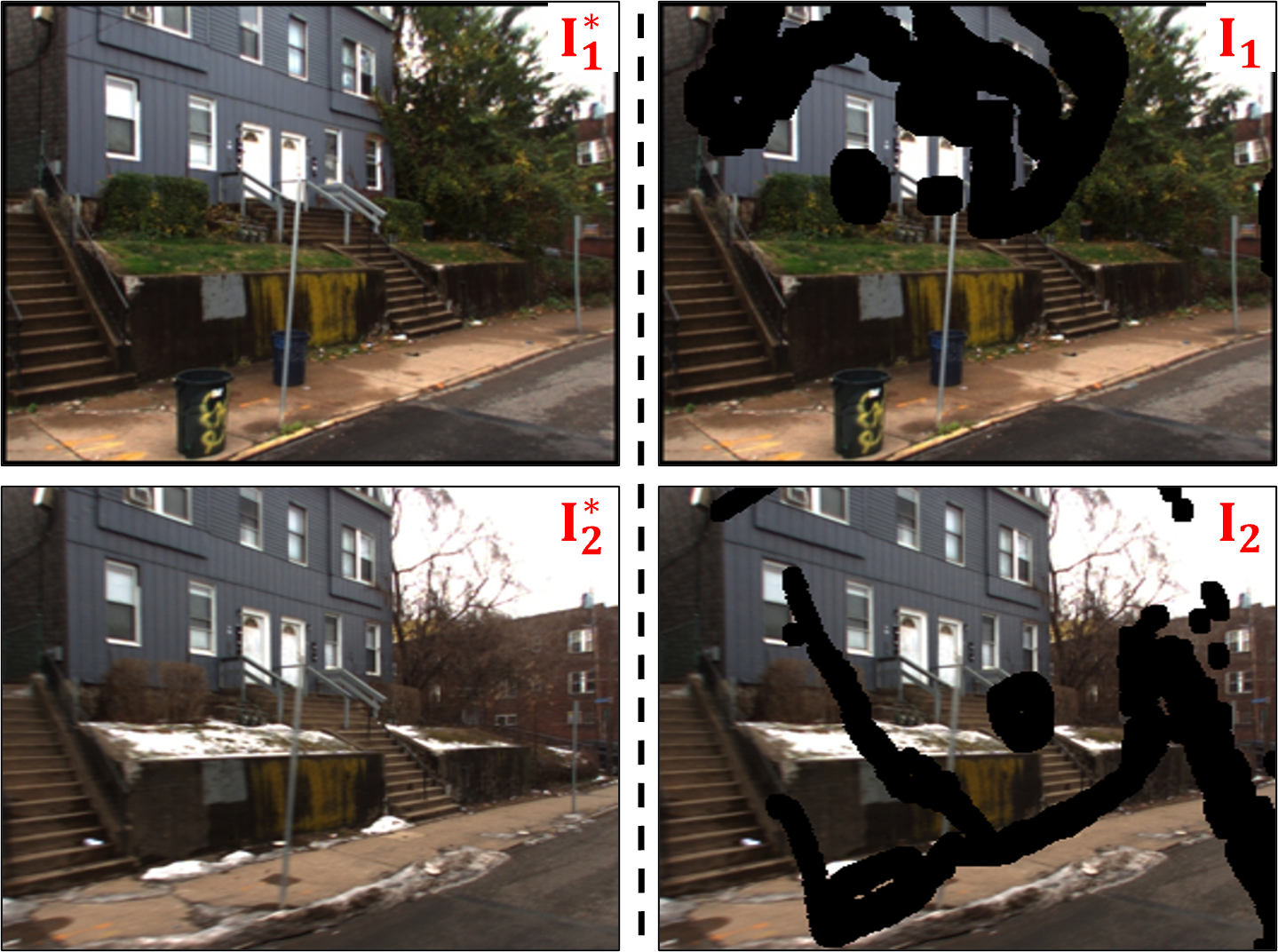}
        \vspace{-6pt}
        \captionof{figure}{Illustration of the ground-truth, the target and reference time-variant images for InDiTE boosting experiments.}
        \label{fig:idteboost_gt}
    \end{minipage}
\end{figure*}
\begin{figure*}[ht]
    \centering
    \includegraphics[width=\linewidth]{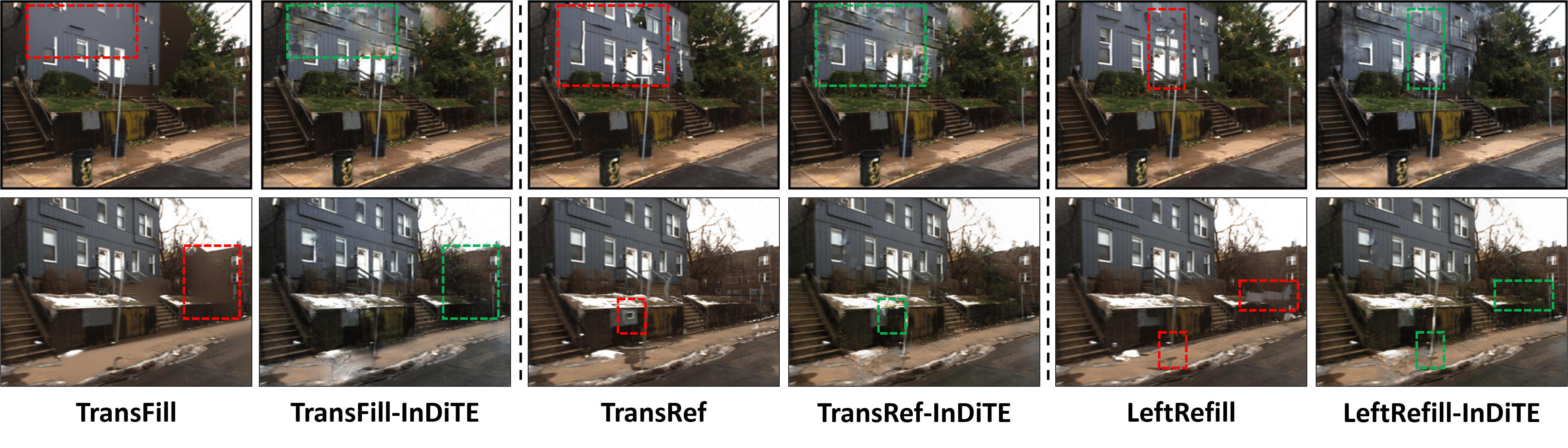}
    \caption{Visualization of inpainting results for the time-variant images in \figref{fig:idteboost_gt} by existing reference-guided image inpainting methods \emph{v.s.} corresponding InDiTE-boosted version. The defects and improvements are highlighted with \textcolor{red}{red} and \textcolor{green}{green} rectangles respectively.}
    \label{fig:idteboost}
\end{figure*}

\subsection{Time-variant Duo-image Inpainting}
We first summarize the overall performance for tvDuoInpaint in \tableref{tab:tamp_result}, then present typical visualization of the results in \figref{fig:tamp_vis} for a more comprehensive experimental analysis. 

\subsubsection{Quantitative Analysis.}
As the results tabulated in \tableref{tab:tamp_result}, we have following observations by comparing our InDiTE-Diff with three baselines for solving tvDuoInpaint task.
\ding{182} Compare to the most competitive baseline, \ie, LeftRefill, our ITDiff generally achieves better results under all the mask ratios for TAMP-Street.
Especially, our InDiTE-Diff outperforms LeftRefill with a large margin, \eg, 3.095\%/3.4442\%, 2.8607\%/2.7819\%, 3.0685\%/3.7822\%, 5.2252\%/5.0317\% PSNR improvements under the four mask ratios respectively.
While the SSIM metric dose not consistently outperforms LeftRefill, we see our method keeps its competitiveness by maintaining the second best results with a very limited performance deficiency.
\ding{183} Moreover, we notice that there is a huge performance gap between the tvDuoInpaint and tvRefInpaint results, which highlights the critical importance of the reference image quality.
However, as we have discussed in \secref{sec:intro}, an intact reference image is not always accessible during real practice.
\tableref{tab:tamp_result} demonstrates the serious of such a fact where existing methods cannot realize comparable tvDuoInpaint results to tvRefInpaint even under the mild mask ratio.
In summary, tvDuoInpaint exposes severe defects of existing methods while our InDiTE-Diff can achieve better performance but more advanced methodology are still needed.

\subsubsection{Qualitative Analysis.}
To have a more clear comparison of the tvDuoInpaint results, we show the comparative visualization results of conventional reference-guided image inpainting methods and our InDiTE-Diff in \figref{fig:tamp_vis}.
In specific, we showcase three different time-variant situations for fair comparison.
It can be observed that our InDiTE-Diff realizes the most reasonable results when compared with existing SOTA baselines, where the deficiencies are highlighted with \textcolor{red}{red} rectangles.
Such visual quality boosting is much clearer when comparing our InDiTE-Diff results with traditional methods' outputs, \ie, TransFill.

\subsection{Boosting Existing Methods with InDiTE}
\label{ssec:boost}
As the proposed InDiTE outputs pairs of (image, mask) which are suitable for any further inpainting process, thus here we further apply it to existing methods to prove its superior complementation capability.
As the PSNR results tabulated in \tableref{tab:itdeboost}, we evaluate both baselines and the corresponding InDiTE boosted version under tvRefInpaint and tvDuoInpaint setup.
It is clear that our InDiTE generally boosted existing methods for time-variant image inpainting tasks. 
Notably, our InDiTE consistently improves the performance for all mask ratios for tvDuoInpaint which proves its great image interactive complementation capability.
While the improvements are not consistent for tvRefInpaint, we note that the reference image is intact under tvRefInpaint setup which already capable of providing high quality complementation.

To have a clear comparison, we also showcase the inpainting results of existing methods when integrated with InDiTE.
\figref{fig:idteboost_gt} shows the tested intact and damaged time-variant images with 20\%-30\% mask ratio. 
Note that we select the image pairs without significant geometric misalignment for exclusive InDiTE boosting verification.
%
As \figref{fig:idteboost} illustrated, we can see that all three methods can only generate blurred or cluttered contents for large masked regions, \ie, the red rectangles.
With InDiTE boosting, all of those regions can restore the original semantic structure as indicated in green rectangles.
While the details are still insufficient, we attribute it to the failure of existing methods for fully utilizing InDiTE outputs which we provide evidence in the next discussion.

%

\begin{figure*}[ht]
    \centering
    \includegraphics[width=\textwidth]{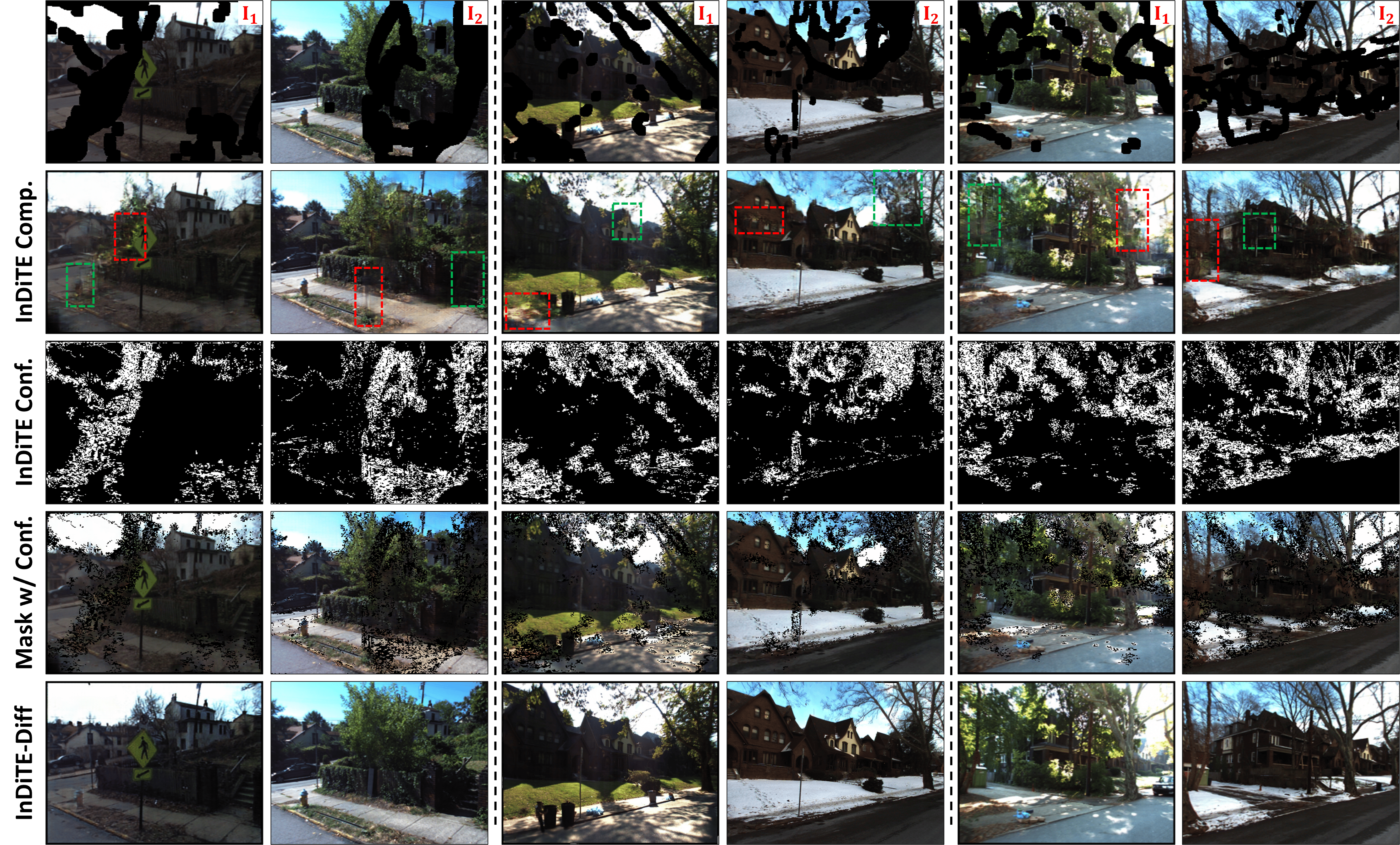}
    \caption{Visualization of InDiTE and InTiDE-Diff results for three exemplar time-variant image pairs. The confidence masks are inversed for better visualization.}
    \label{fig:discuss}
\end{figure*}

\subsection{Discussion}
\label{subsec:discuss}

Throughout the preceding experiments, we did not explicitly elucidate how the proposed InDiTE managed to address the object discrepancy and appearance mismatch challenges that highlighted in the problem formulation, \ie, \secref{subsec:formulate}. 
Indeed, we observe that both the challenges essentially boil down to the same objective: mitigating the semantic discrepancy during image complement, \ie, the insight derived at the beginning of \secref{sec:method}.
In this section, we demonstrate that our InDiTE can achieve such desired image complementation by visualizing the interactive complementation and contradictory content suppression capability of InDiTE in \figref{fig:discuss}.

\subsubsection{Interactive Complementation Capability}
As the InDiTE complemented results shown in \figref{fig:discuss} (\ie, ``InDiTE Comp.'' row), we see that the proposed InDiTE is capable of complementing the time-variant images with information from the counterpart image even both images are damaged.
In particular, it is clear that our InDiTE does not introduce significant semantic contradictions during the complementation.
And after the complementation, the geometric structure of the damaged regions are restored, \ie, the green rectangles.
However, due to the strong semantic consistency constraints the detail textures are smoothed (\ie, enclosed by red rectangles) which are undesirable for the final results.
We show in the next that such detail missing artifacts are further suppressed by utilizing the learned confidence map for further inpainting processing.

\subsubsection{Contradictory Content Suppression}
By learning the confidence of InDiTE results with \reqref{eq:cof_head}, the resulting confidence mask provides indication of whether we should trust the pixel-wise complemented results.
Essentially, such explicit indications support us for realizing the further suppression of content contradiction.
As shown in the ``ITDE Conf.'' row in \figref{fig:discuss}, it can be observed that the resulting confidence maps are highly correlated to the undesirable regions in ``InDiTE Comp.'' and refined the original image masks.
Therefore, when we apply such confidence mask to the InDiTE complemented results, \ie, ``Conf. Masked'' row, we see clearly that those undesired contents in ``ITDE Comp.'' results are further suppressed.
With the further utilization of the generative method, we can achieve much more reasonable results as shown in the ``ITDiff Results'' row in \figref{fig:discuss}. 

In summary, our proposed InDiTE-Diff addresses the time-variant image inpainting challenges with progressive complementation and refinement, where removing either of them will greatly degrade the final performance.
This also explains why the experiments of InDiTE boosting in \secref{ssec:boost} still looks unpromising as shown in \figref{fig:idteboost}.

\begin{table}[t]
    \caption{Ablation results of proposed InDiTE-Diff to solve tvDuoInpaint task on TAMP-Street dataset under 20\%-40\% and 40\%-60\% mask ratios.}
    \label{tab:ablate}
    \resizebox{\linewidth}{!}{
    \begin{tabular}{c|cc|cc}
        \toprule
        \diagbox{Method}{Mask}  & 20\%-40\% & 40\%-60\%  & 20\%-40\% & 40\%-60\% \cr
        \bottomrule
        \toprule
             & \multicolumn{2}{c|}{PSNR $\uparrow$} & \multicolumn{2}{c}{$L_1$ $\downarrow$} \cr
        \cmidrule{1-5}
        DDNM                 & 22.8717/23.7006 & 18.9081/19.0504 & 0.0427/0.0386 & 0.0928/0.0921 \cr 
        DDNM-Interact        & 22.9135/23.7882 & 19.5523/19.6538 & 0.0401/0.0325 & 0.0891/0.0882 \cr 
        InDiTE               & 23.1619/22.9153 & 20.0487/20.1040 & 0.0253/0.0223 & 0.0414/0.0474 \cr
        InDiTE-DDNM          & 23.0512/23.8046 & 19.4225/19.9832 & 0.0260/0.0229 & 0.0929/0.0889 \cr
        \textbf{InDiTE-Diff} & \textbf{24.1764/24.9189} & \textbf{20.5648/21.1180} & \textbf{0.0220/0.0211} & \textbf{0.0413/0.0424} \cr
        \cmidrule{1-5}
            & \multicolumn{2}{c|}{SSIM $\uparrow$} & \multicolumn{2}{c}{LPIPS $\downarrow$} \cr
        \cmidrule{1-5}
        DDNM                 & 0.8755/0.8814 & 0.7526/0.7615 & 0.1493/0.1623 & 0.2832/0.3048 \cr
        DDNM-Interact        & 0.8613/0.8732 & 0.7501/0.7593 & 0.1477/0.1583 & 0.2701/0.2900 \cr 
        InDiTE               & 0.8632/0.8683 & 0.7376/0.7499 & 0.1217/0.1368 & 0.2195/0.2294 \cr
        InDiTE-DDNM          & 0.8627/0.8657 & 0.7323/0.7450 & 0.1222/0.1365 & 0.2304/0.2478 \cr
        \textbf{InDiTE-Diff} & \textbf{0.8953/0.8984} & \textbf{0.7878/0.7801} & \textbf{0.1117/0.1267} & \textbf{0.2192/0.2291} \cr
        \bottomrule
    \end{tabular}
    }
\end{table}

\subsection{Ablation Study}
\label{subsec:ablation}
In this section, we ablate the proposed InDiTE-Diff to verify the contribution of each proposed components.
In specific, we setup four variants for experiments.
\ding{182} We first evaluate DDNM itself as the comparison basis, then \ding{183} setup DDNM with diffusion interaction as \reqref{eq:interact} (\ie, DDNM-Interact) to verify the effectiveness of adding cross-reference during diffusion sampling.
\ding{184} Next, we independently evaluate InDiTE without any further processing, \ie, InTiDE-DDNM.
\ding{185} The final setup is InDiTE-DDNM where the function of \reqref{eq:interact} is removed from InTiDE-Diff.
The ablation results are tabulated in \tableref{tab:ablate}.
In summary, we observe that InDiTE boosts the proposed InDiTE-Diff most under both 20\%-40\% and 40\%-60\% mask ratios.
While the performance lifting of diffusion interaction is limited, it still makes contribute and plays a critical role for achieving the final performance.
%
%
In conclusion, both the design of InDiTE and the cross-reference during diffusion contribute to the overall time-variant image inpainting performance.



\section{Conclusion}

We present the Time-vAriant iMage inPainting (TAMP) task to develop existing image inpainting research by considering the situation where two images are captured with a large time gap, \ie, tvRefInpaint and tvDuoInpaint.
We conduct intuitive experiments and find that even SOTA inpainting methods cannot achieve convincible results due to the failure of proper image complementation.
Thus we design an explicit image complement module, \ie, Interactive Distribution Transition Estimation (InDiTE), for realizing effective image complementation process. 
With the proper complemented results from InDiTE, we further propose the Interactive Distribution Transition Estimation-driven Diffusion (InDiTE-Diff) as our final solution for TAMP task.
Furthermore, considering the lack of benchmarks for TAMP task, we built the TAMP-Street dataset to evaluate the models with fairness. 
The experimental results on TAMP-Street dataset under both tvRefInpaint and tvDuoInpaint setup demonstrate that our proposed model can generally boost existing methods for solving TAMP task.
{
\tiny
\bibliographystyle{IEEEtran}
\bibliography{IEEEabrv, IEEE-Transactions-LaTeX2e-templates-and-instructions/ref}
}

 





\clearpage

\begin{figure*}[t]
    \centering
    \includegraphics[width=\textwidth]{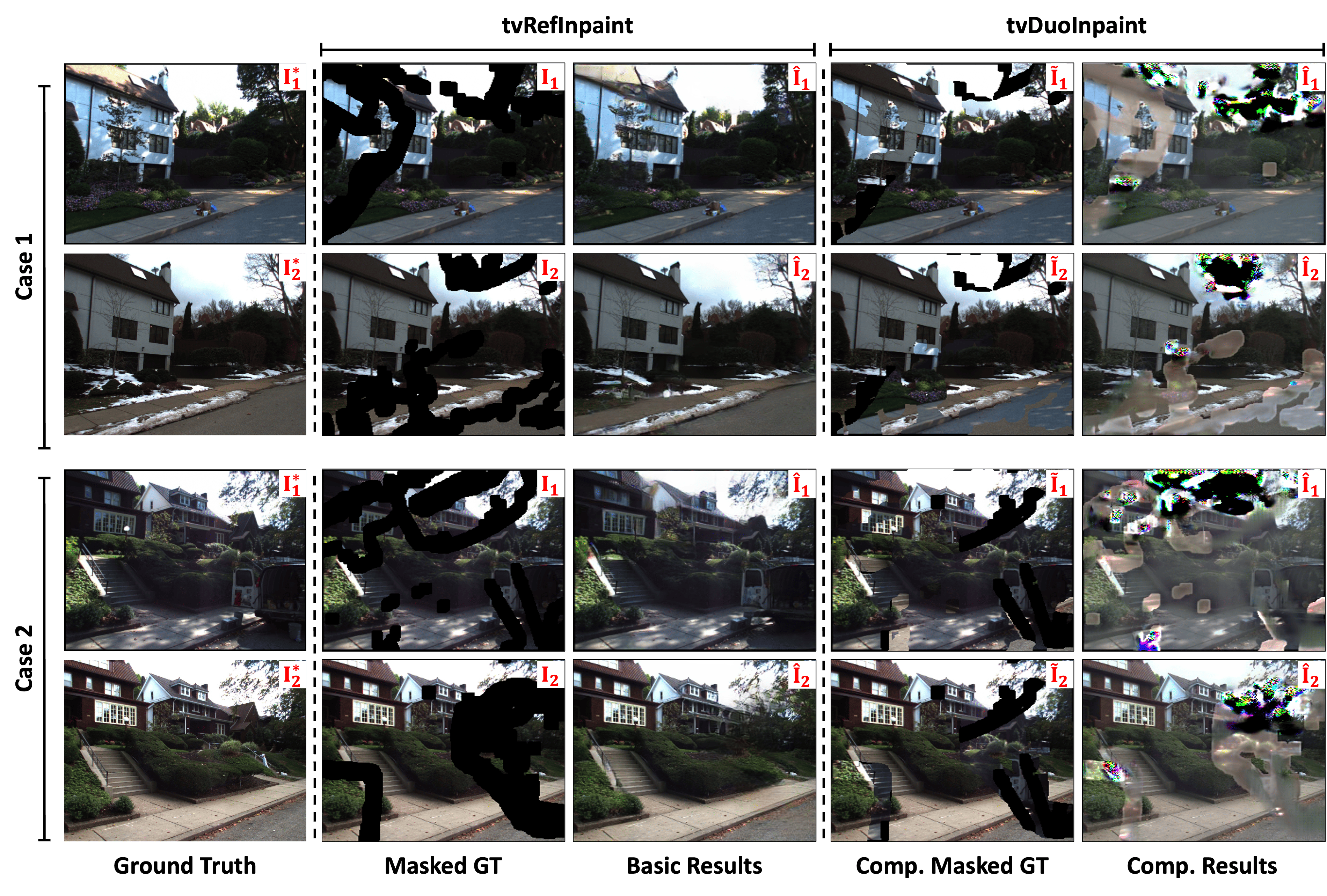}
    \caption{Visualization of another two more LeftRefill inpainting results under tvDuoInpaint setup for intuitive experiments.}
    \label{fig:two-naive}
\end{figure*}

\section*{Intuitive Experiment Details}
\label{app:intuitivexp}

During the intuitive experiments, the official released code for LeftRefill~\footnote{\url{https://github.com/ewrfcas/LeftRefill}} is utilized. We follow the official instruction to train the model on TAMP-Street training split. 
Moreover, we keep resizing the input into 512x512 during the intuitive experiment which is the size used during original model pretraining of LeftRefill.
The model output is further resized to the original image size for final visualization.
For all the other model and training setups of intuitive experiments, we keep them the same as the officially released code.
To better illustrate the naive complementation during intuitive experiment, we have provided the visualization of the procedure in \figref{fig:naive_comp}.
As a supplement for the empirical study, two more time-variant image results under the intuitive experiment setting are shown in Fig.~\ref{fig:two-naive}.

\section*{Small Mask Ratio Results}
\label{app:sratio}

\begin{figure*}[ht]
    \centering
    \includegraphics[width=\textwidth]{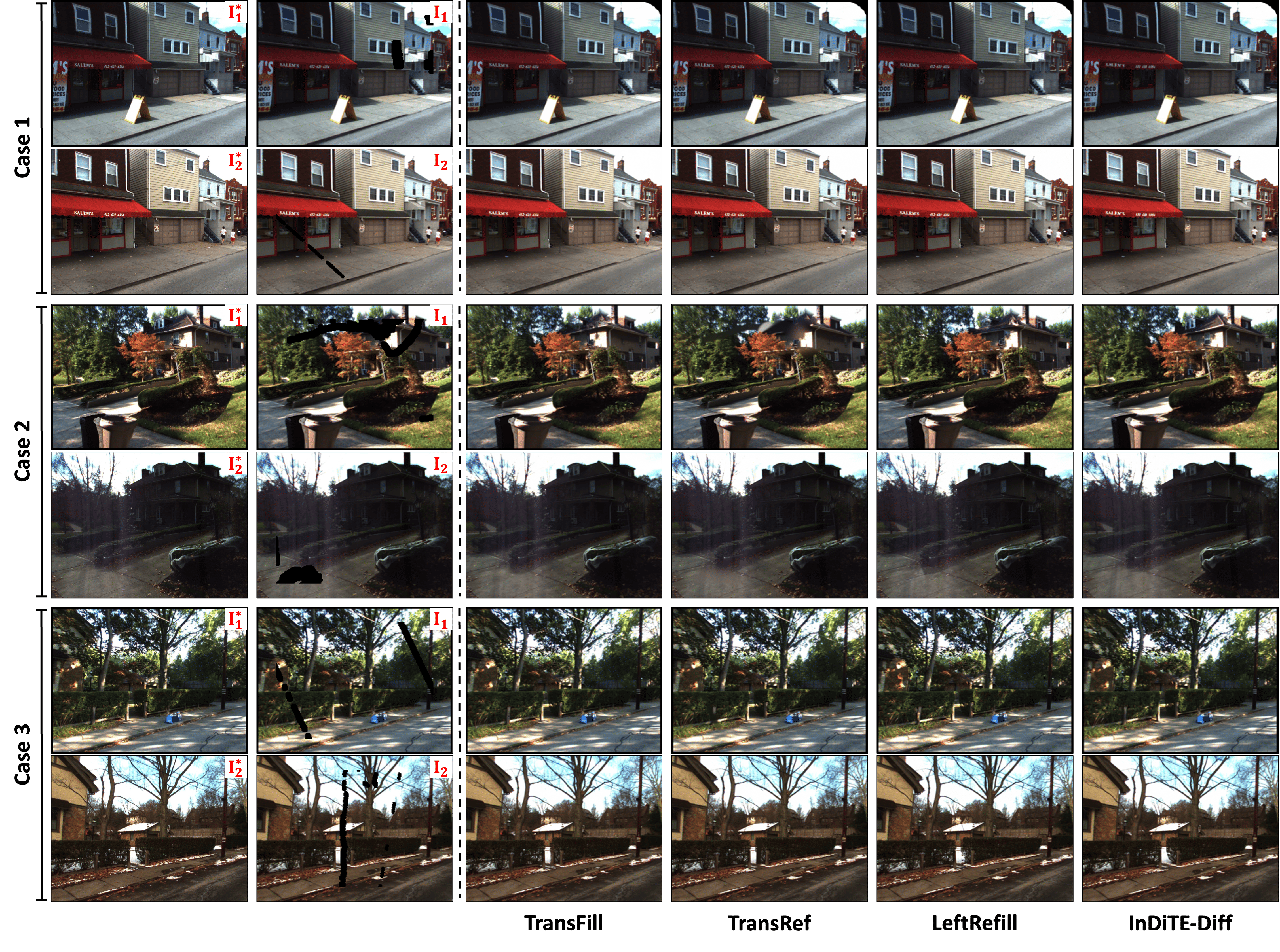}
    \caption{Illustration of SOTA reference-guided image inpainting methods and our InDiTE for inpainting target duo-images with small (\ie, 0\%-20\%) mask ratio.}
    \label{fig:sratio}
\end{figure*}

As shown in Fig.~\ref{fig:sratio}, we test three different cases of small mask ratios for existing reference-guided image inpainting baselines and our InDiTE-Diff.
It is clear that existing diffusion-based image inpainting methods are fully capable of realizing good results when the masked image region is small.
Therefore, as we clarified in Sec.~\ref{sec:data}, we only involve the masks with 20\%-60\% mask ratios to assemble the TAMP-Street dataset.

\section*{Limitations}
\label{app:limit}

As the time-variant images are highly prevalent in daily practice, studying the corresponding inpainting task is of great value for real-world image inpainting applications.
In this work, we defined two variants of time-variant image inpainting task while there still great volume of problems under-resolved.

\paragraph{Dataset.} 
At present, there is no dedicated benchmarks for evaluating TAMP task for which we assembled the TAMP-Street dataset. 
However, TAMP-Street dataset is relatively small in scale when compared with the benchmarks for other machine learning research tasks. 
Moreover, TAMP-Street mainly focuses on real-world urban street situations but there are many other challenging scenarios.
As a result, it would make TAMP more influential and practical by assembling large-scale datasets with sub-datasets for various real-world scenarios. 

\paragraph{Damage Setups.}
In general, the tvDuoInpaint task focuses on the situation where both target and reference images need to be restored at the same time.
However, as one can expect, it is also of great practical application value by considering different image damage situations.
For example, the two images suffer from different degree of pixel missing or both of them involves totally random degree of damage.
Moreover, the pixel damages can also be extended to common corruptions~\cite{hendrycks2019benchmarking} which could expose more vulnerabilities of existing models and necessitate new research efforts.

\section*{Border Impact}

Our research introduces a novel approach to inpainting temporally misaligned images of the same scene, offering promising applications in digital heritage preservation, long-term environmental monitoring, and forensic analysis. By leveraging the shared semantic structure between temporally distant images, this method enables reconstruction even in cases where traditional inpainting fails due to extensive structural changes or degradation. Such capability could assist in restoring historical artifacts or archival media, filling in missing or damaged content by referencing a more recent or older state of the same scene. Furthermore, it opens up possibilities for advanced scene understanding in contexts where change over time is significant, such as disaster recovery, urban development analysis, or climate change documentation.

\section*{Ethical Consideration}

While the technique presents powerful reconstruction capabilities, it also raises ethical concerns around authenticity and misuse. The ability to synthesize or restore parts of an image based on temporally distant data could be exploited to manipulate historical or evidential records, potentially misleading viewers or falsifying context. This is especially critical in domains like journalism, legal evidence, or cultural heritage, where the integrity of visual content is paramount. To mitigate these risks, it is essential to incorporate mechanisms for provenance tracking, transparency about the use of generative reconstruction, and to promote responsible deployment guidelines. Future work should also consider watermarking or metadata embedding to indicate synthesized regions and maintain trust in image authenticity.

\end{document}